\documentclass[a4paper,times]{cjeart}
\usepackage{natbib}
\usepackage{amsthm} 

\graphicspath{{figures/}}
\usepackage{algorithm,algorithmic}

\volumeyear{2026}
\volumenumber{35}
\issuenumber{5}
\journalname{Chinese Journal of Electronics}
\startpage{1}
\endpage{24}
\DOI{10.23919/cje.2025.00.438}
\journaltype{Review}

\usepackage{xspace}
\makeatletter
\DeclareRobustCommand\onedot{\futurelet\@let@token\@onedot}
\def\@onedot{\ifx\@let@token.\else.\null\fi\xspace}
\def\eg{\emph{e.g}\onedot} 
\def\ie{\emph{i.e}\onedot} 
 
\def\etc{\emph{etc}\onedot}

\usepackage{multirow}
\usepackage{makecell}

\usepackage{array, tabularx, boldline}
\usepackage{graphicx}
\usepackage{cellspace}
\usepackage{threeparttable}
\usepackage{bm}

\usepackage{orcidlink} 

\usepackage[capitalize]{cleveref}
\crefname{section}{Sec.}{Secs.}
\Crefname{section}{Section}{Sections}
\Crefname{table}{Table}{Tables}
\crefname{table}{Tab.}{Tabs.}
\crefname{algorithm}{Algo.}{Algos.}

\usepackage[edges]{forest}

\usepackage[edges]{forest}
\newsavebox{\measurebox}
\definecolor{paired-light-blue}{RGB}{198, 219, 239}
\definecolor{paired-dark-blue}{RGB}{49, 130, 188}
\definecolor{paired-light-orange}{RGB}{251, 208, 162}
\definecolor{paired-dark-orange}{RGB}{230, 85, 12}
\definecolor{paired-light-green}{RGB}{199, 233, 193}
\definecolor{paired-dark-green}{RGB}{49, 163, 83}
\definecolor{paired-light-purple}{RGB}{218, 218, 235}
\definecolor{paired-dark-purple}{RGB}{117, 107, 176}
\definecolor{paired-light-gray}{RGB}{217, 217, 217}
\definecolor{paired-dark-gray}{RGB}{99, 99, 99}
\definecolor{paired-light-pink}{RGB}{222, 158, 214}
\definecolor{paired-dark-pink}{RGB}{123, 65, 115}
\definecolor{paired-light-red}{RGB}{231, 150, 156}
\definecolor{paired-dark-red}{RGB}{131, 60, 56}
\definecolor{paired-light-yellow}{RGB}{231, 204, 100}
\definecolor{paired-dark-yellow}{RGB}{141, 209, 49}
\tikzset{%
    parent/.style =          {align=center,text width=1.4cm,rounded corners=3pt, line width=0.3mm, fill=pink!10,draw=pink!80},
    child/.style =           {align=center,text width=2.3cm,rounded corners=3pt, fill=blue!10,draw=blue!80,line width=0.3mm},
    grandchild/.style =      {align=center,text width=2cm,rounded corners=3pt},
    greatgrandchild/.style = {align=center,text width=1.5cm,rounded corners=3pt},
    greatgrandchild2/.style = {align=center,text width=1.5cm,rounded corners=3pt},    
    referenceblock/.style =  {align=center,text width=1.5cm,rounded corners=2pt},
    data/.style =           {align=center,text width=2cm,rounded corners=3pt, fill=paired-light-blue!50,draw=paired-dark-blue!65,line width=0.3mm},
    data_wide/.style =           {align=left,text width=4.1cm,rounded corners=3pt, fill=paired-light-blue!50,draw=paired-dark-blue!65,line width=0.3mm},   
    data_work/.style =           {align=center, text width=4.5cm,rounded corners=3pt, fill=paired-light-blue!50,draw=blue!0,line width=0.3mm},  
    data_work_left/.style =      {align=left, text width=4.45cm,rounded corners=3pt, fill=paired-light-blue!50,draw=paired-dark-blue!65,line width=0.3mm},  
    data_work_small/.style =     {align=left, text width=3cm,rounded corners=3pt, fill=paired-light-blue!50,draw=blue!0,line width=0.3mm},  
    model/.style =           {align=center,text width=2cm,rounded corners=3pt, fill=paired-light-orange!50,draw=paired-dark-orange!65,line width=0.3mm},  
    model_small/.style =           {align=center,text width=1.96cm,rounded corners=3pt, fill=paired-light-orange!50,draw=paired-dark-orange!65,line width=0.3mm}, 
    model_wide/.style =           {align=left,text width=4.1cm,rounded corners=3pt, fill= paired-light-orange!50,draw=paired-dark-orange!65,line width=0.3mm}, 
    model_more/.style =           {align=center,text width=4cm,rounded corners=3pt, fill=paired-light-orange!50,draw=paired-dark-orange!65,line width=0.3mm},
    model_more_left/.style =      {align=left,text width=4.45cm,rounded corners=3pt, fill=paired-light-orange!50,draw=paired-dark-orange!65,line width=0.3mm},
    model_large_left/.style =      {align=left,text width=6.45cm,rounded corners=3pt, fill=paired-light-orange!50,draw=paired-dark-orange!65,line width=0.3mm},   
    model_work/.style =           {align=center,text width=4.5cm,rounded corners=3pt, fill=paired-light-orange!50,draw=red!0,line width=0.3mm},
    model_work_left/.style =      {align=left,text width=4cm,rounded corners=3pt, fill=paired-light-orange!50,draw=red!0,line width=0.3mm}, 
    model_work_small/.style =     {align=left,text width=3cm,rounded corners=3pt, fill=paired-light-orange!50,draw=red!0,line width=0.3mm},  
    model_work_small_2/.style =     {align=left,text width=4cm,rounded corners=3pt, fill=paired-light-orange!50,draw=red!0,line width=0.3mm}, 
    pretraining/.style =           {align=center,text width=2cm,rounded corners=3pt, fill= paired-light-purple!50,draw=paired-dark-purple!75,line width=0.3mm}, 
    pretraining_wide/.style =           {align=center,text width=3cm,rounded corners=3pt, fill= paired-light-purple!50,draw=paired-dark-purple!75,line width=0.3mm}, 
    pretraining_more/.style =           {align=center,text width=4.1cm,rounded corners=3pt, fill=paired-light-purple!50,draw=paired-dark-purple!75,line width=0.3mm},   
    pretraining_work/.style =           {align=center,text width=4.5cm,rounded corners=3pt, fill= paired-light-purple!50,draw= cyan!0,line width=0.3mm},      
    finetuning/.style =           {align=center,text width=2cm,rounded corners=3pt, fill= paired-light-green!50,draw=paired-dark-green!75,line width=0.3mm},   
    finetuning_wide/.style =      {align=left,text width=4.1cm,rounded corners=3pt, fill=paired-light-green!50,draw=paired-dark-green!75,line width=0.3mm},   
    finetuning_work/.style =      {align=center,text width=4.5cm,rounded corners=3pt, fill=paired-light-green!50,draw= orange!0,line width=0.3mm},      
    finetuning_work_left/.style =      {align=left,text width=4.45cm,rounded corners=3pt, fill=paired-light-green!50,draw= paired-dark-green!75,line width=0.3mm},
    finetuning_work_small/.style =     {align=left,text width=3cm,rounded corners=3pt, fill=paired-light-green!50,draw= orange!0,line width=0.3mm},
    inference/.style =           {align=center,text width=2cm,rounded corners=3pt, fill= paired-light-red!35,draw=paired-light-red!90,line width=0.3mm},           
    inference_more/.style =      {align=center,text width=4.45cm,rounded corners=3pt, fill= paired-light-red!35,draw=paired-light-red!90,line width=0.3mm},
    inference_work/.style =      {align=left,text width=4cm,rounded corners=3pt, fill= paired-light-red!35,draw= magenta!0,line width=0.3mm},      
    inference_work_long/.style =      {align=left,text width=9cm,rounded corners=3pt, fill= paired-light-red!35,draw= paired-light-red!90,line width=0.3mm},      
    application/.style =           {align=center,text width=2cm,rounded corners=3pt, fill= paired-light-yellow!15,draw=paired-light-yellow!90,line width=0.3mm},           
    application_more/.style =      {align=center,text width=11.4cm,rounded corners=3pt, fill= paired-light-yellow!15,draw=paired-light-yellow!90,line width=0.3mm},
    application_more_left/.style =      {align=left,text width=9cm,rounded corners=3pt, fill= paired-light-yellow!15,draw=paired-light-yellow!90,line width=0.3mm},
    application_work/.style =      {align=center,text width=4.5cm,rounded corners=3pt, fill= paired-light-yellow!15,draw= magenta!0,line width=0.3mm},   
}

\title[A Survey on the Green Development of Large Models: From Resource-Efficient Architectures to Hardware-Software Co-Design]{A Survey on the Green Development of Large Models: From Resource-Efficient Architectures to Hardware-Software Co-Design}

\author{%
Linhui Xiao\affilnums{1}$^{\orcidlink{0000-0003-2592-5264}}$, 
Guiping Cao\affilnums{1}$^{\orcidlink{0000-0002-0682-2158}}$, 
Mingyue Guo\affilnums{1}$^{\orcidlink{0009-0005-2348-7530}}$,
Xianchao Guan\affilnums{1,2}$^{\orcidlink{0009-0003-1384-0527}}$, 
Fan Yang\affilnums{1}$^{\orcidlink{0009-0005-0123-3505}}$, 
Ming Tao\affilnums{1}$^{\orcidlink{0000-0002-4662-7170}}$,\\ 
Xin Li\affilnums{1,*}$^{\orcidlink{0000-0002-1670-1368}}$,
Yuxin Peng\affilnums{3}$^{\orcidlink{0000-0001-7658-3845}}$, 
Yaowei Wang\affilnums{2,1,*}$^{\orcidlink{0000-0002-6110-4036}}$
}

\affiliation{%
\affilnum{1}Pengcheng Laboratory, Shenzhen 518066, China \\
\affilnum{2}Harbin Institute of Technology (Shenzhen), Shenzhen 518055, China \\
\affilnum{3}Peking University, Beijing 100080, China \\
}

\corrauth{Xin Li, Yaowei Wang}
\email{xinlihitsz@gmail.com, wangyw@pcl.ac.cn}

\receivetime{Manuscript Received September 30, 2025}
\accepttime{Accepted January 13, 2026}
\publishtime{Published Online February 7, 2026}

\authorcopy{Firstname1 Middlename1 Lastname1\emph{et al.}}

\abstract{
The rapid expansion of large-scale AI models has led to significant performance breakthroughs across diverse domains, yet it has also raised critical concerns regarding computational costs, energy consumption, and environmental sustainability. This survey provides a comprehensive overview of the green development of large models, emphasizing resource-efficient architectures and full-stack hardware-software co-design. We systematically review recent advances in efficient model construction, including attention operator optimization, linear-complexity architectures, and model sparsification and merging, as well as training and deployment strategies such as data-efficient learning, parameter-efficient fine-tuning, and computational compression. Beyond algorithmic improvements, we explore energy-efficient AI hardware, including mainstream AI chips, memory optimization, cross-platform deployment, and sustainable infrastructure. Furthermore, we examine how large models are being applied to sustainability-critical domains such as DeepSeek, remote sensing interpretation, national-scale infrastructure, and global initiatives. Finally, we discuss key challenges and future directions, highlighting the need for continual learning paradigms, memory-centric hardware, and standardized evaluation protocols. This survey aims to offer a holistic roadmap toward sustainable, scalable, and socially responsible development of large models.
}


\begin{document}

\maketitle

\section{Introduction}
\label{sec:chapter1}

In recent years, large-scale Artificial Intelligence (AI) models, especially those built upon Transformer architectures~\cite{vaswani2017attention}, have achieved significant breakthroughs in natural language processing, computer vision, and scientific computing. Flagship models such as BERT~\cite{devlin2019bert}, CLIP~\cite{radford2021learning}, LLaVA~\cite{lin2024moe}, and the GPT series~\cite{openai2024gpt4technicalreport} have continuously expanded the frontier of AI. However, this progress has come with substantial costs: the exponential increase in model parameters, training data, and input length has led to skyrocketing computational demands, resulting in massive energy consumption and environmental concerns, as shown in Table \ref{tab:chap1_green}. As model parameters scale from billions to trillions and input sequences extend from 1K to 100K tokens, training and inference costs have increased superlinearly, posing major challenges to the \textit{accessibility}, \textit{scalability}, and \textit{sustainability} of large-scale AI systems.\footnote{\textcolor{blue}{https://cje.ejournal.org.cn/article/doi/10.23919/cje.2025.00.438}}

\begin{figure}[t!]
\vspace{8pt}
\centering
\includegraphics[width=0.39\textwidth]{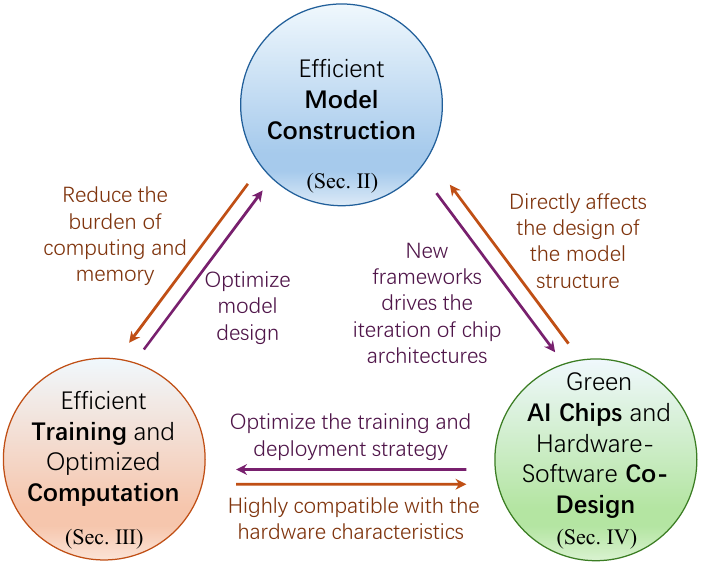}
\vspace{-2pt}
\caption{A triangular framework illustrating the layered relationship among model architecture, training strategies, and green AI hardware.}
\label{fig:chap1_framework}
\vspace{-12pt}
\end{figure}

\begin{table}[t!]
\setlength\tabcolsep{4pt}
\caption{Comparison of GPU hours, Energy Usage, and Carbon Footprint for training mainstream large models. $^\dag$ indicates no official data available, and the values are estimated and for reference only.}
\label{tab:chap1_green}
\resizebox{1.0\linewidth}{!}{
 \renewcommand\arraystretch{1.2}
\begin{tabular}{c|c|c|c}
\hline
Model & \makecell{GPU hours \\ (h)}&  \makecell{Eneregy Usage \\(KWh)}  & \makecell{Carbon Emitted \\ (tCO$_2$eq) }    \\ \hline
CLIP & 256k V100 & 71k & 24$^\dag$ \\ \hline
BLOOM & 1,083k A100 & 475k & 183 \\ \hline
GPT-3 & 3,140k$\sim$4,600k$^\dag$ V100 & 1,287k$^\dag$ & 552$^\dag$ \\ \hline
GPT-4o & $\sim$60,000k$^\dag$ H100 & 16,800k$^\dag$ & 21,660$^\dag$\\ \hline
Gemini 1 & 40,000k$\sim$50,000k$^\dag$ TPUv4 & 8,000k$\sim$12,000k$^\dag$ & 4,000$\sim$6,000$^\dag$ \\ \hline
Gemini 2 & 12,000k$\sim$15,000k$^\dag$ TPUv6 & 3,000k$\sim$5,000k$^\dag$ & 1,500$\sim$2,500$^\dag$ \\ \hline
Qwen3-Max & $\sim$300,000k$^\dag$ H200 & 23,000k$^\dag$ & - \\ \hline
DeepSeek-V3 & 2,788k H800 & 1,087k$^\dag$  & 584 \\ \hline
LLAMA & 1,022k A100 & 449k  & 173 \\ \hline
LLAMA 2 & 1,720k A100 & 688k$^\dag$  & 291 \\ \hline
LLAMA 3 & 30,840k H100 & $>$11,000k  & 11,390 \\ \hline
SAM & 18k A100 & 7k  & 3 \\ \hline  
SAM 3 & 172k A100 or 86k H200 & 142k$\sim$176k  & 66$\sim$78 \\
\hline  
\end{tabular}}
\centering
\vspace{-10pt}
\end{table}

The core of this inefficiency lies in the architectural design of dominant models. The attention mechanism \cite{vaswani2017attention} in Transformers scales with $O(N^2)$ time and memory complexity, creating the well-known “\textit{quadratic wall}” that severely limits the processing of long contexts. Moreover, the dense activation paradigm requires every parameter in the model to participate in each computation, irrespective of its relevance, resulting in substantial waste of computational resources and memory. These foundational limitations have made the training of leading-edge models extremely resource-intensive: for instance, training a 100K-context model can incur nearly 100$\times$ the cost of training a 10K-context model. The environmental impact is equally concerning, with recent studies estimating that training a single large model can emit as much carbon as the lifetime emissions of multiple cars \cite{radovanovic2022carbon}.

Numerous surveys have emerged in response to these challenges~\cite{guo2025survey, duan2024efficient, shen2024efficient, bai2024beyond}. However, most of them \cite{guo2025survey, duan2024efficient, shen2024efficient} concentrate on isolated technical components, such as Parameter-Efficient Fine-Tuning (PEFT), quantization, or pruning, without providing a system-level perspective. In particular, they often neglect the role of hardware constraints and co-design strategies, and typically lack coverage of emerging techniques like physics-inspired models, state-space modeling, or model merging. Furthermore, prior works rarely explore real-world sustainability applications, such as remote sensing or climate modeling, where efficient AI can make a tangible environmental impact.

To address these challenges systematically, this survey adopts a top-down perspective that reflects both the logical hierarchy and structural dependencies of efficient AI systems. As illustrated in Figure~\ref{fig:chap1_framework}, we organize the \textbf{green development of large models} into a \textbf{triangular framework} spanning three tightly coupled layers: \textbf{(1)} the top layer focuses on resource-efficient architectures, where innovations in model design shape computational patterns and compatibility with downstream hardware; \textbf{(2)} the middle layer emphasizes training and deployment optimization, bridging algorithmic techniques with system constraints; and \textbf{(3)} the bottom layer centers on hardware–software co-design, where energy-efficient chips, inference memory optimization, and collaborative cross-platform deployment support upper-layer models. This structure is not merely hierarchical, but also bidirectional: architectural breakthroughs influence chip design, while hardware capabilities inform model structure and optimization strategy. Unlike prior literature, we present a unified view of how efficiency gains emerge from layer-wise collaboration rather than isolated techniques.

\begin{figure*}
\centering
\scriptsize
\hspace*{-0pt}
    \begin{forest}
        for tree={
            forked edges,
            grow'=0,
            draw,
            rounded corners,
            node options={align=center,},
            text width=2.7cm,
            s sep=3pt,  
            calign=edge midpoint,
        },
        [\textbf{Survey} \\ \textbf{on the} \\ \textbf{Green} \\ \textbf{Development} \\ \textbf{of Large Models} , fill=gray!45, parent
            [\S \ref{sec:chapter1} Introduction, for tree={pretraining}
                [Research Status and Our Pipeline, pretraining_more]
            ]
            [\S \ref{sec:chapter2} Efficient \\ Model \\ Construction, for tree={data}
                [\S \ref{subsec:operator_optimization} Attention Operator Optimization, data_wide
                    [\S \ref{subsubsec:conventional_self_attention} Quadratic Complexity Analysis, data_work_left
                        [CLIP~\cite{radford2021learning}; GPT~\cite{openai2024gpt4technicalreport}; \etc, data_work_small]
                    ]
                    [\S \ref{subsubsec:linear_attention} Linear Attention, data_work_left
                        [Linear~\cite{katharopoulos2020transformers}; \etc, data_work_small]
                    ]                
                    [\S \ref{subsubsec:flash_attention} Flash Attention, data_work_left
                        [FlashAttention~\cite{dao2022flashattention,dao2023flashattention,shah2024flashattention}; \etc, data_work_small]
                    ]
                ]
                [\S \ref{subsec:model_framework_design} Efficient Model
                Design,  data_wide  
                    [\S \ref{subsubsec:physics_inspired_models_vheat} Physics-Inspired Models, data_work_left
                        [vHeat~\cite{wang2025building}; \etc, data_work_small]
                    ]
                    [\S \ref{subsubsec:state_space_models_ssms} State Space Models, data_work_left
                        [VMamba~\cite{liu2024vmamba}; \etc, data_work_small]
                    ]
                    [\S \ref{subsubsec:other_linear_complexity_models} Other Linear-Complexity Models, data_work_left
                        [RWKV~\cite{peng2023rwkv}; RetNet~\cite{sun2023retentive}, data_work_small]
                    ]
                ]
                [\S \ref{subsec:structure_optimization} Model Sparsification and Merging, data_wide
                    [\S \ref{subsubsec:sparsely_activated_models} Sparse Activation Mechanism, data_work_left
                        [MoE~\cite{jacobs1991textordfeminineadaptive, shazeer2017outrageously, lepikhin2020gshard} ; \etc, data_work_small]
                    ]
                    [\S \ref{subsubsec:model_merging} Model Merging, data_work_left
                        [Survey~\cite{enneng2024Merging}; \etc, data_work_small]
                    ]
                ]
            ]
            [\S\ref{sec:chapter3} Efficient Training and Optimized Computation, for tree={fill=red!45,model}
                [\S \ref{subsec:data_efficient_techniques} Data-Efficient Techniques, model_wide
                    [\S \ref{subsubsec:data_efficient_sampling} Data-Efficient Sampling, model_more_left
                    [SAI-DPO~\cite{rao2025dynamic}; \etc, model_work_small]
                    ]
                    [\S \ref{subsubsec:data_efficient_distillation} Data-Efficient Distillation, model_more_left
                    [KRR~\cite{nguyendataset,zhou2022dataset,loo2022efficient}; \etc, model_work_small]
                    ]
                    [\S \ref{subsubsec:data_efficient_pruning} Data-Efficient Pruning, model_more_left
                        [DeepCore\cite{guo2022deepcore}; \etc, model_work_small]
                    ]
                ]
                [\S \ref{subsec:distributed_training_techniques} Distributed Training Techniques, model_wide
                    [\S \ref{subsubsec:data_and_model_parallelism} Data and Model Parallelism, model_more_left  
                        [\cite{li2020pytorch,rajbhandari2020zero,rasley2020deepspeed}; ~\cite{shoeybi2019megatron,huang2019gpipe,narayanan2021efficient}; \etc, model_work_small]
                    ]
                    [\S \ref{subsubsec:mixed_precision_training} Mix-Precision Training, model_more_left
                        [BLOOM~\cite{workshop2022bloom}; \etc, model_work_small]
                    ]
                ]
                [\S \ref{subsec:parameter_efficient_fine_tuning} Parameter-Efficient Fine-Tuning, model_wide
                    [\S \ref{subsubsec:prompt-based_tuning} Prompt-based Tuning, model_more_left
                        [prompt ensembling~\cite{lester2021power}; \etc, model_work_small]
                    ]
                    [\S \ref{subsubsec:adapter_tuning} Adapter Tuning, model_more_left
                        [Adapter \cite{houlsby2019parameter}; \etc, model_work_small]
                    ]
                    [\S \ref{subsubsec:low_rank_adaptation} Low-Rank Adaption ,model_more_left
                        [vanilla LoRA \cite{hu2022lora}; \etc, model_work_small]
                    ]
                ]
                [\S \ref{subsec:model_compression} Model and Computational \\  ~~~~~~Compression, model_wide
                    [\S \ref{subsubsec:low_precision_quantization}  Low-Precision Quantization, model_more_left
                        [LLM-QAT \cite{liu2023llm};  
                        \etc, model_work_small]
                    ]
                    [\S \ref{subsubsec:model_pruning} Model Pruning, model_more_left
                        [SparseGPT \cite{frantar2023sparsegpt}; 
                        \etc, model_work_small]
                    ]
                    [\S \ref{subsubsec:knowledge_distillation} Knowledge Distillation, model_more_left
                        [TinyLLM \cite{kandala2024tinyllm};
                        \etc, model_work_small]
                    ]
                    [\S \ref{subsubsec:speculative_decoding} Speculative Decoding, model_more_left
                        [SpecEE \cite{xu2025specee}; \etc, model_work_small]
                    ]
                ]
            ]
            [\S \ref{sec:chapter4} Green \\ Computing AI Chips and Hardware-Software Co-Design, for tree={finetuning}
                [\S \ref{subsec:mainstream_ai_chips_and_application_ecology} Mainstream AI Chips and \\ ~~~~~~Application Ecology, finetuning_wide
                    [\S \ref{subsubsec:classification_and_comparison_of_ai_chips} Classification of AI Chips, finetuning_work_left
                    [Survey~\cite{hu2022survey}; \etc, finetuning_work_small]
                    ]
                    [\S \ref{subsubsec:global_ai_chip_ecosystems} Global AI Chip Ecosystems, finetuning_work_left
                    [TPU~\cite{jouppi2017datacenter}; \etc, finetuning_work_small]
                    ]
                    [\S \ref{subsubsec:domestic_ai_chip_development_ecosystems} Domestic AI Chip Development, finetuning_work_left
                    [Cambricon~\cite{liu2016cambricon}; \etc, finetuning_work_small]
                    ]
                ]
                [\S \ref{subsec:inference_memory_optimization} Inference Memory Optimization, finetuning_wide
                    [\S \ref{subsubsec:kv_cache_optimization} KV Cache Optimization, finetuning_work_left
                        [LaCache \cite{shi2025lacache}; 
                         \etc, finetuning_work_small]
                    ]
                    [\S \ref{subsubsec:memory_offloading}  Memory Offloading, finetuning_work_left
                        [Aqua \cite{vijaya2025aqua}; HeadInfer \cite{luo2025headinfer}, finetuning_work_small]
                    ]
                    [\S \ref{subsubsec:early_exiting}  Early Exiting, finetuning_work_left
                        [AdaInfer \cite{fan2024not}; 
                        \etc, finetuning_work_small]
                    ]                         
                ]
                [\S \ref{subsec:cross_platform_deployment_and_adaptation} Hardware-Software Collaborative \\  ~~~~~~Cross-platform Deployment, finetuning_wide
                    [\S \ref{subsubsec:unified_inference_framework} Unified Inference Framework, finetuning_work_left
                        [LLMBox \cite{tang2024llmbox}; 
                        \etc, finetuning_work_small]
                    ]
                    [\S \ref{subsubsec:automatic_operator_scheduling} Automatic Operator Scheduling, finetuning_work_left
                        [LightLLM \cite{gong2025past}; \etc, finetuning_work_small]
                    ]
                    [\S \ref{subsubsec:key_operator_discretization} Key Operator Discretization, finetuning_work_left
                        [Qtile \cite{zhang2025qfactory}; 
                        KPerfIR \cite{guan2025kperfir}; \etc, finetuning_work_small]
                    ]
                ]
                [\S \ref{subsec:energy_efficient_hardware_system} Energy-Efficient Hardware System, finetuning_wide
                    [\S \ref{subsubsec:emerging_green_chip_technologies} Emerging Green Chip Technologies, finetuning_work_left
                        [neuromorphic chip~\cite{basu2022spiking}; \etc, finetuning_work_small]
                    ]
                    [\S \ref{subsubsec:decentralized_green_learning} Decentralized Green Learning, finetuning_work_left
                        [PETALS~\cite{borzunov2023petals}; \etc, finetuning_work_small]
                    ]
                    [\S \ref{subsubsec:green_data_centers} Green Data Centers, finetuning_work_left
                        [\cite{cao2023data,radovanovic2022carbon,zhang2025decarbonizing}; \etc, finetuning_work_small]
                    ]
                ]
            ]
            [\S \ref{sec:chapter5} AI for \\ Sustainability Application, for tree={application}
                [\S \ref{subsec:deeepseek} Parallel Training and RL-Driven Reasoning Paradigm, application_more_left]
                [\S \ref{subsec:remote_sensing_interpretation} Remote Sensing Interpretation: RS-vHeat and “Aerospace·Lingmou” 3.0, application_more_left]
                [\S \ref{subsec:china_computing_net} China Computing NET (C$^2$NET): A National-Scale Sustainable Infrastructure, application_more_left]
                [\S \ref{subsec:google_meta_sustainable_ai} Google and Meta: Full-Stack Optimization for Sustainable AI, application_more_left]
                [\S \ref{subsec:global_ai_applications_for_sustainability} Global AI Applications for Multi-Domain Sustainability, application_more_left]
            ]
            [\S \ref{sec:chapter6}  Discussion and Future Outlook, for tree={inference}
                [\S \ref{subsec:learning_withot_starting_over} Learning without Starting Over: Continual; Incremental; and Federated Training, inference_work_long
                ]
                [\S \ref{subsec:co_designing_models_with_memory_and_physics_centric_compute} Co-Designing Models with Memory and Physics-Centric Compute, inference_work_long
                ]
                [\S \ref{subsec:edge_ai} Edge AI: Efficiency on Constrained Devices, inference_work_long
                ]
                [\S \ref{subsec:standardized_and_unified_evaluation} Standardized and Unified Evaluation, inference_work_long
                ]
            ]
            [\S \ref{sec:chapter7} Conclusion, for tree={pretraining}
                [Summary of the Full Text, pretraining_more]
            ]
        ]
    \end{forest}
    \vspace{-2pt}
    \caption{Overview of the paper structure, detailing Chapter \ref{sec:chapter1}-\ref{sec:chapter7}.}
    \vspace{-15pt}
    \label{fig:paper_structure}
\end{figure*}
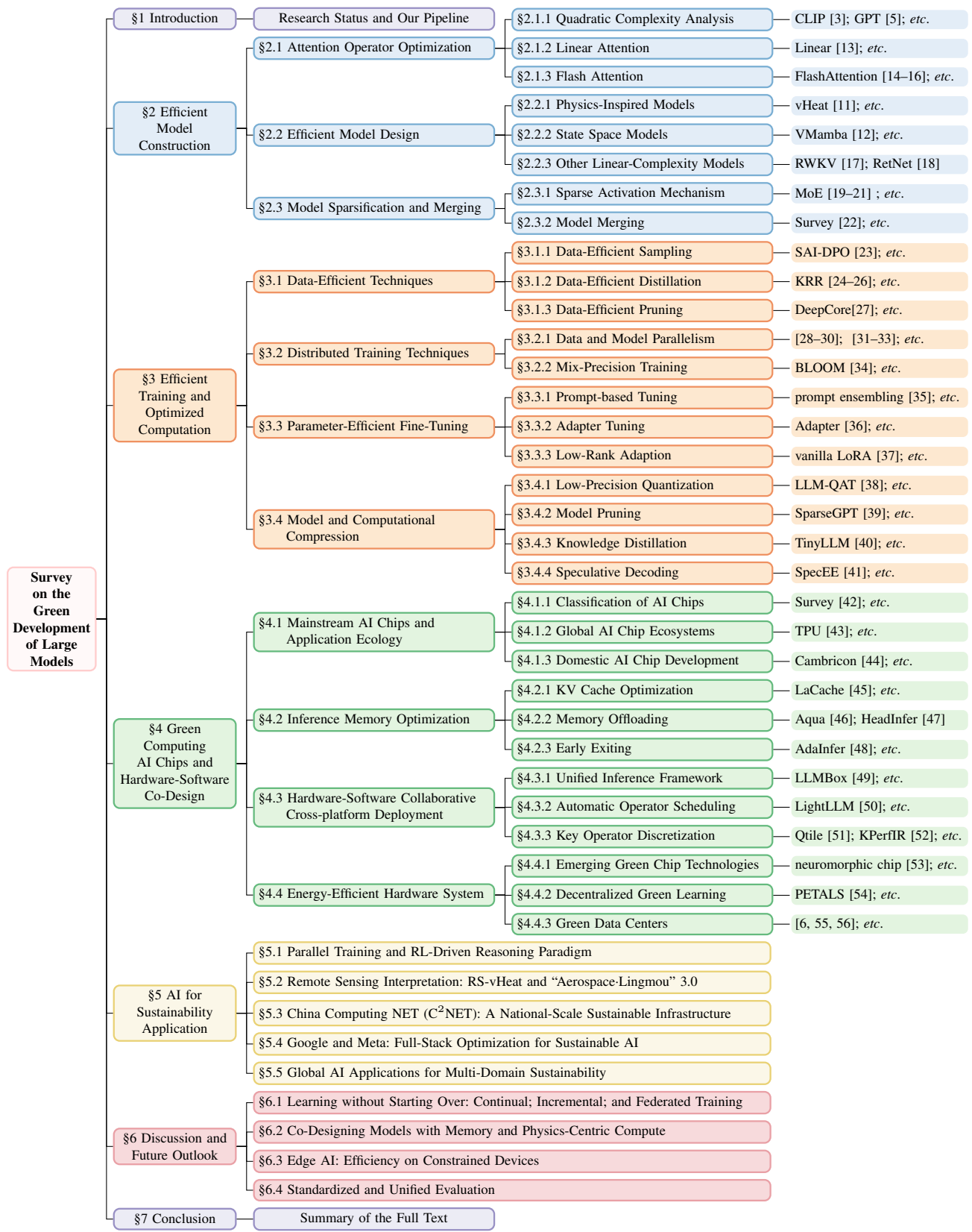

As shown in Figure~\ref{fig:paper_structure}, following this framework, \textbf{\cref{sec:chapter2}} surveys architectural-level innovations, including physics-inspired models like vHeat~\cite{wang2025building}, state-space models such as VMamba~\cite{liu2024vmamba}, sparsity techniques like Mixture of Experts (MoE), and model merging. \textbf{\cref{sec:chapter3}} focuses on training and optimized computation, covering data-efficient learning strategies, distributed training, parameter-efficient fine-tuning, model and computational compression, such as quantization, distillation, and speculative decoding. \textbf{\cref{sec:chapter4}} examines the green AI hardware, highlighting memory optimization, cross-platform deployment, and energy-aware hardware system design. \textbf{\cref{sec:chapter5}} extends the discussion to AI for sustainability, covering recent Chinese advances like DeepSeek and the “Aerospace·Lingmou” 3.0 system, national-scale infrastructure such as C$^2$NET, and global initiatives led by Google, Meta, and others. Finally, \textbf{\cref{sec:chapter6}} discusses future opportunities, including continual learning, neuromorphic computing, edge AI, and the pressing need for standardized efficiency benchmarks. Collectively, these chapters provide a layered and interconnected perspective on the emerging landscape of green AI.

By integrating model design, algorithm optimization, and hardware implementation into a unified framework, this survey aims to provide both theoretical insights and practical guidance for developing scalable, low-carbon AI systems. We emphasize the interplay of sparsity, modularity, and hardware-awareness as recurring principles across layers, and advocate for a lifecycle-aware approach to AI sustainability, from pretraining to deployment and beyond.

In summary, this survey contributes to the field by offering a unified, full-stack perspective that emphasizes the interdependence between model design, optimization strategies, and hardware implementation. We introduce a triangular framework that captures the bidirectional influences among layers and systematically organizes recent advances within it. Additionally, we highlight how design decisions at one layer affect constraints and opportunities at others. By synthesizing a broad range of techniques into a cohesive view, we aim to support researchers and practitioners in navigating the complex trade-offs involved in building efficient, scalable, and sustainable large-scale AI systems.

\section{Efficient Model Construction}
\label{sec:chapter2}
The rapid progress in large-scale neural networks has driven remarkable advances in AI, but at the expense of sharply increasing computational and memory costs. As models and context lengths grow, the conventional Transformer architecture encounters fundamental scalability issues, particularly due to the quadratic complexity of its attention mechanism, which becomes a major bottleneck in long-sequence scenarios. This section provides a review of cutting-edge innovations at the architectural level that are designed to mitigate the computational and memory demands of large models. We focus on three major research directions: (1) \textit{attention operator optimization}; (2) \textit{efficient model design}; (3) \textit{model sparsification and merging}. Each of these approaches seeks to decouple model performance from resource consumption, opening pathways to sustainable and scalable AI development.

\subsection{Attention Operator Optimization}
\label{subsec:operator_optimization}

Computational operators are the core kernels executing tensor operations, a function crucial for enhancing computational efficiency.
A representative case is the attention mechanism in Transformers, whose quadratic complexity with respect to sequence length introduces fundamental bottlenecks in memory and speed.
To overcome these limitations, research has pivoted to hardware-aware and efficiency-driven operator designs. The following section highlights several representative operator designs.

%
%
\subsubsection{\textbf{Quadratic Complexity in Attention Mechanism}}
\label{subsubsec:conventional_self_attention}

Traditional attention mechanisms achieve global dependency modeling by computing correlations between every position in a sequence through the scaled dot-product attention formula: 
$\operatorname{Attention}(\bm{Q}, \bm{K}, \bm{V})=\operatorname{softmax}\left(\frac{\bm{Q} \bm{K}^T}{\sqrt{d_k}}\right) \bm{V}$
%
, where $\bm{Q}$ (queries), $\bm{K}$ (keys), and $\bm{V}$ (values) are linear projections of the input sequence. 
This process involves calculating $\bm{Q} \bm{K}^T$ to generate an $N \times N$ similarity matrix, scaling it, applying softmax for normalization, and finally weighting $\bm{V}$ vectors. Multi-head self-/cross-attention extends this by running multiple attention functions in parallel to capture diverse representations. However, the core limitation arises from the $\bm{Q} \bm{K}^T$ computation, which exhibits quadratic complexity $O\left(N^2\right)$ in both computation and memory requirements. For example, when sequence length increases from 1K to 32K tokens, the attention matrix storage demand grows by a factor of 1,024, creating a severe ``\textit{quadratic wall}'' that hinders long-sequence processing. Despite the success of models like BERT~\cite{devlin2019bert}, CLIP~\cite{radford2021learning}, and GPT~\cite{liu2024gpt} in NLP and vision tasks, this complexity poses significant challenges for practical deployment in scenarios requiring long-context understanding, such as long document analysis, high-resolution image processing, and extended video sequences.
\subsubsection{\textbf{Linear Attention}}
\label{subsubsec:linear_attention}
Following the quadratic complexity challenges of traditional Transformers, Linear Attention~\cite{katharopoulos2020transformers} reduces computational complexity from $O\left(N^2\right)$ to linear $O(N)$ through feature mapping functions $\phi(\bm{Q})$ and $\phi(\bm{K})$ that transform Query and Key matrices into higher-dimensional feature representations. The key insight reformulates attention computation by applying feature mappings such as $\phi(x)=\operatorname{ReLU}(x)$ or $\phi(x)=\operatorname{ELU}(x)+1$ to queries and keys, then leveraging matrix multiplication associativity to reorganize computation order and avoid explicit attention matrix construction. Instead of computing the full $\bm{Q} \bm{K}^T$ attention matrix, Linear Attention precomputes the combinations of mapped keys with values and reuses them for every query, achieving linear complexity. While this approach theoretically enables processing of arbitrarily long sequences, Linear Attention suffers from significant performance degradation, particularly in tasks requiring precise positional information and local dependency modeling, as the linearization process loses the sharp, focused distributions that traditional Softmax attention provides. Representative works include Linformer~\cite{wang2020linformer} and Linear Transformer~\cite{schlag2021linear}, primarily applied to long document understanding and time series prediction, but practical deployment remains limited due to substantial accuracy losses that often cannot be justified by efficiency gains in core NLP tasks.

\subsubsection{\textbf{Flash Attention}}
\label{subsubsec:flash_attention}


Flash Attention~\cite{dao2022flashattention} addresses the memory bottleneck of standard attention mechanisms through tiled computation and memory access optimization, maintaining $O\left(N^2\right)$ computational complexity while achieving significant memory efficiency improvements. Unlike Linear Attention's complexity reduction approach, Flash Attention reorganizes computation patterns by decomposing attention into blocks that fit within GPU's SRAM, using online softmax algorithms to avoid materializing complete attention matrices and reducing memory complexity from $O\left(N^2\right)$ to $O(N)$. 
The technique minimizes expensive memory transfers between HBM (High Bandwidth Memory) and SRAM, achieving 2-4$\times$ speedups for sequences up to 8K tokens, though computational costs for ultra-long sequences ($>$100$\mathrm{K}$ tokens) remain prohibitive due to the unchanged algorithmic complexity. The Flash Attention family has evolved through v1/v2/v3 iterations~\cite{dao2022flashattention, dao2023flashattention, shah2024flashattention}, with v2 adding variable sequence length support and v3 enhancing hardware utilization, becoming ubiquitous in production systems including GPT-4~\cite{openai2024gpt4technicalreport}, Claude~\cite{anthropic2024claude}, and LLaMA~\cite{touvron2023llama}. This memory optimization approach is complementary to Linear Attention's complexity reduction, suggesting potential for hybrid methods that combine both algorithmic efficiency and implementation optimizations for next-generation attention mechanisms.

These operator-level optimizations enable longer sequences and lower latency while bridging the gap between theoretical algorithm efficiency and real-world performance. Together, they form the foundation of modern deep learning engines, allowing full-stack computation redesign without changing model semantics. Such innovations are essential for efficient training and deployment of next-generation large models, making AI computing more efficient and accessible.

\begin{figure*}[h!]
\centering
\includegraphics[width=1\textwidth]{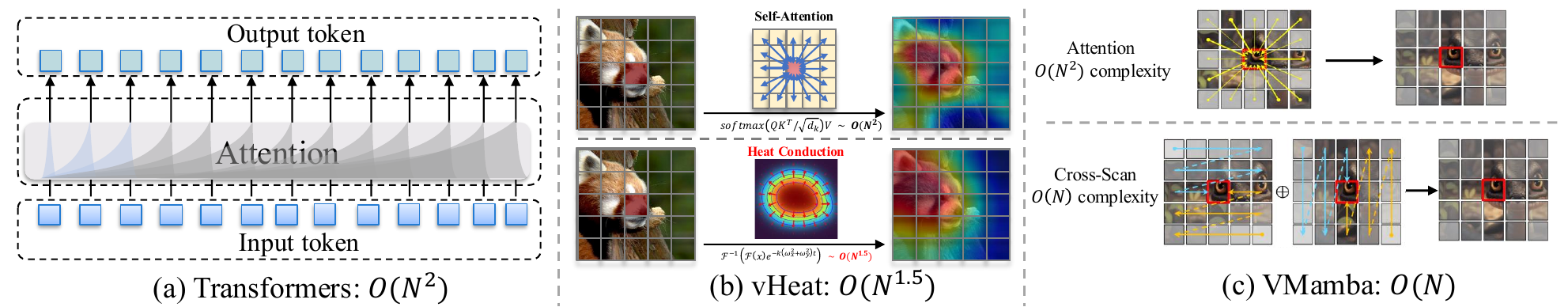}
\vspace{-18pt}
\caption{Computational complexity comparison of fundamental architectures: (a) Transformers \cite{vaswani2017attention} with $O(N^2)$ self-attention, (b) vHeat \cite{wang2025building} with $O(N^{1.5})$ heat conduction operator, and (c) VMamba \cite{liu2024vmamba} with $O(N)$ cross-scan mechanism.}
\label{architecture}
\vspace{-10pt}
\end{figure*}

\subsection{Efficient Model Design}
\label{subsec:model_framework_design}

To advance the computational efficiency of large-scale models, it is essential to not only optimize computational operators for enhancing microscopic tensor computations but also adopt a holistic approach toward efficient model construction that improves end-to-end model efficiency.

The following sections explore a range of efficient models and the innovative techniques behind them, highlighting their collective contribution to improved efficiency.

\subsubsection{\textbf{Physics-Inspired Models}}
\label{subsubsec:physics_inspired_models_vheat}
\begin{figure}[h!]
\centering
\includegraphics[width=0.48\textwidth]{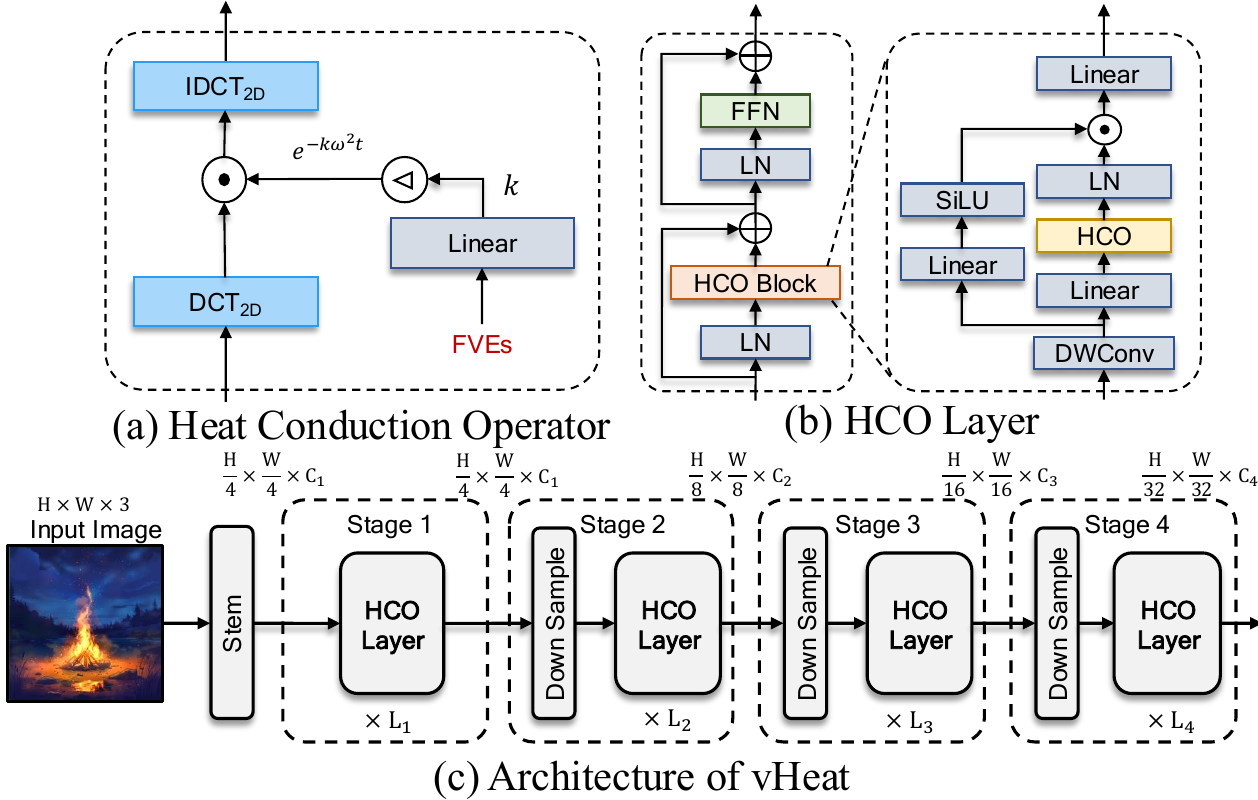}
\vspace{-6pt}
\caption{Illustration of vHeat's \cite{wang2025building} framework. It shows the heat conduction-inspired mechanism and the integration of the Discrete Cosine Transform for efficient global dependency modeling.}
\vspace{-9pt}
\label{vHeat2}
\end{figure}

The vHeat~\cite{wang2025building} model represents innovative exploration in physics-inspired modeling, achieving complexity reduction from $O\left(N^2\right)$ to $O\left(N^{1.5}\right)$ by analogizing visual information propagation to heat conduction phenomena. 
As illustrated in Figure~\ref{vHeat2}, the model's core innovation introduces the Heat Conduction Operator (HCO), designed based on general solutions of 2D heat conduction equations, simulating heat diffusion processes in the frequency domain through efficient Discrete Cosine Transform (DCT) and Inverse DCT (IDCT). Adaptive heat diffusion coefficients are dynamically predicted through learnable Frequency Value Embeddings (FVEs), achieving global receptive fields while maintaining sub-quadratic complexity. vHeat's primary limitation lies in the approximation nature of physical simulation, which may not fully capture complex semantic dependencies. 
The model demonstrates excellent performance in image classification and object detection tasks, with particularly notable advantages in remote sensing image processing~\cite{hu2024rs}. This establishes a critical theoretical and technical foundation for the future development of linear complexity models.

\subsubsection{\textbf{State Space Models}}
\label{subsubsec:state_space_models_ssms}

State Space Models (SSMs) have emerged as a powerful alternative to traditional attention-based architectures, achieving breakthroughs in efficient long-sequence modeling. By introducing state variables to encode historical context, SSMs transform sequence modeling from quadratic attention computation to linear-time state transitions, reducing time and space complexity from $O\left(N^2\right)$ to $O\left(N\right)$.
%
The foundational principle of SSMs lies in their formulation as discrete-time dynamical systems, mathematically governed by state transition matrix $\bm{A}$, input matrix $\bm{B}$, and output matrix $\bm{C}$. This formulation allows the model to capture long-range dependencies through continuous state evolution, avoiding the need for pairwise token interactions.
Representative models include S4~\cite{gu2021efficiently}, Mamba~\cite{gu2023mamba}, and its vision variant VMamba~\cite{liu2024vmamba}.
%
S4~\cite{gu2021efficiently} introduced a hardware-aware efficient algorithm and demonstrated strong performance in long-range reasoning tasks.
%
Mamba~\cite{gu2023mamba} further advanced SSMs by proposing the Selective Scan Mechanism (SSM), which enables state transition parameters to dynamically adapt based on input content. This overcomes a key limitation of earlier SSMs, which struggled with content-aware reasoning.
%
VMamba~\cite{liu2024vmamba} extended these principles to computer vision, introducing a cross-scale scanning strategy to maintain global receptive fields in visual tasks. The architecture achieved a $2.8\times$ speedup and reduced GPU memory usage by $86.8\%$ in high-resolution image processing, showcasing the versatility of state-space methods beyond language.

These successes highlight the role of SSMs in enabling scalable and efficient inference in data-intensive domains, offering a promising path toward sustainable and high-performance AI systems.

\subsubsection{\textbf{Other Linear-Complexity Models}}
\label{subsubsec:other_linear_complexity_models}
Beyond the mainstream SSMs development trajectory, the academic community has explored numerous innovative approaches to linear complexity architectures, each addressing specific aspects of the quadratic bottleneck through distinct technical pathways.

\textbf{RWKV} (Receptance Weighted Key Value)~\cite{peng2023rwkv} achieves linear complexity language modeling through ingenious fusion of RNN's recurrent mechanisms with Transformer's parallel training advantages. Its core innovation reformulates the attention mechanism into recursive form, maintaining global information access capabilities while achieving linear complexity and elegantly bridging the gap between sequential processing efficiency and parallel training scalability.

\textbf{RetNet} (Retentive Network)~\cite{sun2023retentive} proposes the Retention Mechanism, which incorporates relative positional encoding to support parallelization during training while enabling recurrent computation during inference. This design achieves dual optimization of both training and inference efficiency, addressing the fundamental trade-off between training parallelizability and inference efficiency that has long challenged sequence modeling architectures.

As illustrated in Figure~\ref{architecture}, these architectural innovations demonstrate varying levels of complexity reduction, with conventional Self-Attention constrained by $O\left(N^2\right)$ self-attention complexity, while novel approaches like vHeat achieve $O\left(N^{1.5}\right)$ complexity through heat conduction operators, and architectures such as VMamba reach linear $O\left(N\right)$ complexity via cross-scan mechanisms.
These diverse explorations collectively demonstrate the rich technical ecosystem emerging around linear complexity architectures, providing multiple potential pathways for future development and driving efficient sequence modeling technology toward more mature and practical applications.

%
\subsection{\textbf{Model Sparsification and Merging}}
\label{subsec:structure_optimization}
%
Building on advances in attention operators and efficient architectures, we now examine complementary approaches focused on optimizing and consolidating existing models rather than designing new ones from scratch. These strategies reduce the footprint of large models while preserving capabilities, directly addressing computational and memory constraints.
The following sections will delve into the technical intricacies and recent advancements within these domains.

\begin{figure}[t!]
\centering
\includegraphics[width=0.45\textwidth]{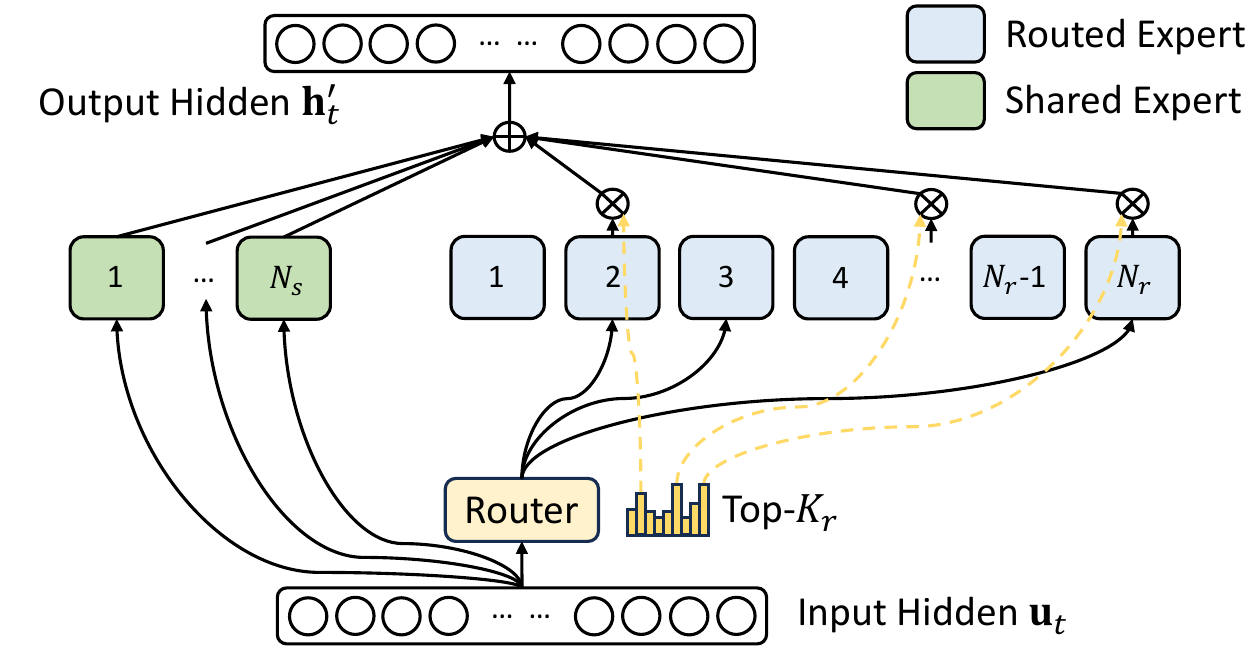}
\vspace{-7pt}
\caption{Illustration of the Mixture of Experts (MoE) architecture~\cite{dai2024deepseekmoe}. MoE employs a gating network to dynamically route input tokens to a subset of specialized expert networks, enabling efficient sparse activation and improved scalability for large-scale models.}
\vspace{-4mm}
\label{MoE}
\end{figure}

\subsubsection{\textbf{Sparse Activation Mechanism}}
\label{subsubsec:sparsely_activated_models}
The core idea of sparse activation is to break through these computational bottlenecks through clever design: each input token activates only a small subset of parameters in the network, dramatically reducing actual computational requirements while maintaining or even improving model performance. 
The Mixture of Experts (MoE) architecture~\cite{jacobs1991textordfeminineadaptive, shazeer2017outrageously, lepikhin2020gshard} represents the most successful implementation of this principle, as shown in Figure~\ref{MoE}, replacing traditional dense FFN layers with multiple specialized expert sub-networks and employing learnable gating mechanisms to determine which experts process each token.

The mathematical foundation of MoE can be expressed precisely: An MoE layer contains $N$ expert networks $[E_1, E_2, \ldots, E_N]$, where a gating network generates logits normalized through softmax distribution. The gating value for expert $i$ is:
\begin{equation}
G(x)=\operatorname{softmax}\left(g_1(x), g_2(x), \ldots, g_N(x)\right)
\end{equation}
Top-k gating selects the experts with highest probabilities, defining active expert set $\tau$, and the MoE output becomes:
\begin{equation}
\operatorname{MoE}(x)=\sum_{i \in \tau} G(x)_i \cdot E_i(x)
\end{equation}
This design decomposes complex problems into specialized subtasks, with each expert handling specific input patterns, achieving dual improvements in computational efficiency and model capability while dramatically reducing computation through selective expert activation.

\begin{table*}[]
\caption{Systematic classification of MoE method paradigms. A comprehensive taxonomy of seven distinct MoE architectures spanning from early fixed assignment approaches to modern adaptive systems, highlighting their key characteristics, representative applications, and primary advantages across different deployment scenarios and performance objectives.}
\centering
\resizebox{0.99\linewidth}{!}{
\begin{tabular}{l|c|c|c}
\hline
Paradigm   & Key Characteristics & Main Applications & Advantages \\ \hline
Static MoE  & Fixed expert-input mapping  & MoE~\cite{jacobs1991textordfeminineadaptive} & Early theoretical work \\ 
Dynamic Sparse MoE & Learnable gating, top-k activation & GShard~\cite{lepikhin2020gshard}, Switch Transformer~\cite{shazeer2017outrageously} & Large language models \\ 
Dense MoE & All experts activated & Soft MoE~\cite{puigcerver2023sparse} & High-accuracy scenarios \\ 
Hierarchical MoE & Multi-level expert organization & Deep MoE~\cite{Wang2018dmoe}, PLE~\cite{tang2020progressive} & Complex structured problems \\ 
Multi-task MoE & Task-specific gating & MMoE~\cite{ma2018modeling}, PLE~\cite{tang2020progressive} & Recommendation systems \\ 
Multi-modal MoE & Modality-specific experts & V- MoE~\cite{riquelme2021scaling}, MoE-LLaVA~\cite{lin2024moe} & Vision-language tasks \\ 
Adaptive/Evolutionary MoE & Dynamic expert adjustment & AdaMoE~\cite{zeng2024adamoe}, EvoMoE~\cite{jing2025evomoe} & Research frontiers \\ \hline
\end{tabular}}
\centering
\label{moe_paradigms}
\vspace{-10pt}
\end{table*}

As shown in Table~\ref{moe_paradigms}, the evolution of MoE technology has led to diverse architectural paradigms, each addressing specific computational challenges and application requirements. Based on fundamental differences in routing strategies, activation patterns, and specialization mechanisms, MoE architectures can be categorized into seven distinct paradigms that represent the technical spectrum from early fixed assignment approaches to modern adaptive systems. Each paradigm offers unique advantages for different deployment scenarios and performance objectives.

\vspace{-5pt}
\begin{figure}[h!]
\centering
\includegraphics[width=0.48\textwidth]{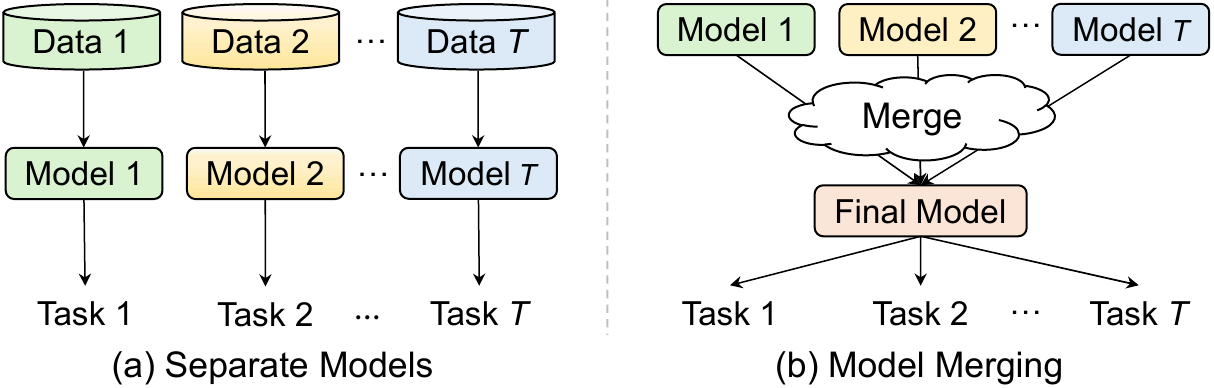}
\vspace{-7pt}
\caption{Illustration of the model merging paradigm~\cite{enneng2024Merging}. (a) $T$ separate models for $T$ tasks, (b) A merged model for $T$ tasks.}
\label{model_merging}
\vspace{-2mm}
\end{figure}

\subsubsection{\textbf{Model Merging}}
\label{subsubsec:model_merging}
This frontier investigates approaches to combine the knowledge and capabilities of multiple pre-trained models, each potentially specialized for different tasks, into a single, unified model without access to the original training data. Known as Model Merging~\cite{enneng2024Merging}, this process aims to construct a generalist model by combining several specialists. As shown in Figure~\ref{model_merging}, it represents a form of zero-shot compression and knowledge integration, eliminating the need to store or deploy multiple separate models and thus reducing both memory overhead and inference latency.

The techniques can be broadly categorized into four key areas each contributing unique advancements: foundational methods like Task Arithmetic~\cite{gabriel2023editing}, which establishes basic principles through task vector operations to combine or remove capabilities; efficient and scalable techniques such as Model Soups~\cite{wortsman2022model}, which employs weight averaging to enhance generalization without increasing inference costs; specialized approaches including ZipIt~\cite{stoica2023zipit}, which tailors merging for heterogeneous architectures through feature alignment; and theoretical frameworks exemplified by Linear Mode Connectivity (LMC)~\cite{entezari2021role,frankle2020linear}, which explains loss landscape connectivity to justify weight interpolation. 

Based on the aforementioned approaches, such as specifically model sparsification and merging, not only effectively tackle scalability challenges in large-scale AI systems but also advance sustainable AI development by encouraging resource-efficient model reuse and collaborative integration. Collectively, they form a robust post-design methodology that produces highly efficient, versatile, and scalable models, ultimately paving the way for more sustainable and adaptable AI ecosystems. These framework-level efforts enhance the overall efficiency of AI deployments while minimizing computational costs and environmental impact.


\vspace{7pt}
\section{Efficient Training and Optimized Computation}
\label{sec:chapter3}

The scaling of large-scale models to billions of parameters creates a severe computational bottleneck, making training and computation exceptionally costly, time-consuming, and environmentally unsustainable due to the substantial carbon footprint. To address these challenges, a number of techniques have been introduced across complementary fronts for developing efficient training, including \textit{data-efficient techniques}, \textit{distributed training techniques}, \textit{parameter-efficient fine-tuning}, and \textit{model and computational compression}. This section details these essential approaches and their impacts for achieving green AI development.






\subsection{\textbf{Data-Efficient Techniques}}
\label{subsec:data_efficient_techniques}

The reliance of large-scale models on massive datasets makes their training computationally intensive and environmentally costly. Data-efficient techniques mitigate this by maximizing the utility of every data sample, reducing resource use without compromising performance.
According to the mode of data operation, this review categorizes data-efficient techniques into three areas: \textit{sampling}, \textit{distillation}, and \textit{pruning}. 
 

\subsubsection{\textbf{Data-Efficient Sampling}}
\label{subsubsec:data_efficient_sampling}

Non-uniform sampling reduces training data volume without compromising model performance by selectively emphasizing the most informative examples. As a static data scheduling strategy used in LLM pre-training (\eg, GPT-3~\cite{brown2020language}, LLaMA~\cite{touvron2023llama1}), it assigns manual sampling weights to upsample high-quality data and downsample lower-quality content, thereby increasing exposure to valuable information while preserving diversity.

In contrast to this static approach, Dynamic Data Sampling is a technique that dynamically adapts the sampling strategy throughout the training process based on data characteristics, model state, or task requirements.
These methods are categorized according to their adaptation criteria.

\noindent\textbf{\textit{Difficulty-Based Sampling}}. It prioritizes samples by difficulty, often starting with easier examples to stabilize training before introducing harder ones. Curriculum Learning (CL)~\cite{bengio2009curriculum,wang2021survey,xiao2023clip} trains models on data ordered from easy to hard, mimicking human educational progression. 
Recent extensions, such as SAI-DPO~\cite{rao2025dynamic}, dynamically select training data for mathematical reasoning using ``self-aware difficulty" metrics across different training phases,
thus enhancing both data utilization efficiency and final task performance.
    
\noindent\textbf{\textit{Loss-Based Sampling}}. This approach dynamically prioritizes training examples by their loss values or gradient norms to focus on informative or poorly learned samples. Core methods include direct loss reweighting (\eg, up-weighting high-loss samples), and approximated loss sampling (using proxy models to predict loss and reduce computation). The loss value is introduced as an alternative metric in pioneering approaches~\cite{katharopoulos2017biased,jiang2019accelerating} to construct a sampling distribution that reduces gradient variance compared to uniform sampling. Importance Sampling~\cite{katharopoulos2018not} reduces variance by weighting samples according to a tractable upper bound on the per-sample gradient norm. 
Recent dynamic loss-based method~\cite{sow2025dynamic} employs instance-level data reweighting to prioritize informative samples, emphasizing challenging data while reducing focus on redundant ones. 
    

\subsubsection{\textbf{Data-Efficient Distillation}}
\label{subsubsec:data_efficient_distillation}


Dataset distillation~\cite{wang2018dataset} aims to produce a compressed, highly informative synthetic subset that preserves the core features and distribution of original dataset, thereby drastically reducing its size. A comprehensive review~\cite{yu2023dataset} outlines a general framework for data distillation. The methods are grouped into three categories: \textit{performance matching}, \textit{parameter matching}, and \textit{distribution matching}.

\noindent\textbf{\textit{Performance Matching}}. This method optimizes synthetic data to ensure that models trained on it achieve performance comparable to those trained on original data, as measured by minimal loss on the original dataset. They can be further categorized into two types: (1) Meta learning-based approaches~\cite{wang2018dataset,deng2022remember}, which are computationally expensive and require substantial GPU memory; (2) Kernel Ridge Regression (KRR)-based methods~\cite{nguyendataset,zhou2022dataset,loo2022efficient}, which employ convex optimization and yield a closed-form solution for linear models, avoiding the need for extensive inner-loop training.

\noindent\textbf{\textit{Parameter Matching}}. The core of this approach involves training an identical network architecture separately on both the synthetic and original datasets, while enforcing consistency between the model parameters derived from both data sources. Considering the number of training steps, this approach can be further categorized into single-step parameter matching~\cite{zhang2023accelerating} and multi-step parameter matching~\cite{cazenavette2022dataset,li2024dataset}. Single-step methods focus on computational efficiency by following instantaneous gradients, while multi-step methods aim for higher accuracy by converging to optimal parameter states via multi-step optimization.

\noindent\textbf{\textit{Distribution Matching}}. This approach aims to align the distribution of synthetic data with that of real data. Distribution matching differs from other methods by directly minimizing the distribution distance between synthetic and real data, employing metrics like Maximum Mean Discrepancy (MMD)~\cite{zhao2023dataset}. CAFE~\cite{wang2022cafe} constrains the feature statistics of synthetic and real samples to be consistent across all network layers except the last one.
To better capture distributional differences, NCFD~\cite{wang2025dataset} formulates dataset distillation as a min-max optimization problem using the Neural Characteristic Function Discrepancy, which is a novel and theoretical metric for distribution comparison. 

\subsubsection{\textbf{Data-Efficient Pruning}}
\label{subsubsec:data_efficient_pruning}


Data pruning strives to remove redundant samples to retain the most informative subset of dataset, known as the coreset. Research on data pruning primarily follows two approaches: \textit{score-based} and \textit{geometry-based} methods.

\noindent\textbf{\textit{Score-Based}}. This method assigns importance metrics to data points and retain those with the highest scores. DeepCore~\cite{guo2022deepcore} constructs a comprehensive code library and provide an empirical study on popular coreset selection methods. By calculating the average $L_2$ norm of the error vector, E2LN~\cite{paul2021deep} assesses the importance of training examples, identifying crucial examples very early in training. The field has further evolved to include a variety of strategies, such as those based on ``forgetting events'' to measure how often each example is forgotten during training~\cite{tonevaempirical}, prioritizing uncertain samples based on prediction variation (Dyn-Unc~\cite{he2024large}), and hybrid strategies like TDDS's dual-depth pruning~\cite{zhang2024spanning} that combine difficulty and uncertainty.


\noindent\textbf{\textit{Geometry-Based}}. This approach aims to construct a coreset that better represents the underlying data distribution~\cite{agarwal2020contextual}. To construct representative coresets, various methods have been proposed: Herding~\cite{welling2009herding} minimizes distribution discrepancy; SSP~\cite{sorscher2022beyond} and Moderate~\cite{xia2022moderate} reduce redundancy by selecting distant or median samples, though this may harm generalization by overlooking difficult examples. Addressing this, D2 pruning~\cite{maharanamathbb} uses a graph-based message-passing mechanism to enhance diversity and model generalization.


In brief, data-efficient techniques are shifting from static, rule-based strategies toward adaptive and intelligently scheduled systems. 
The trend is toward intelligent data orchestrators. Powered by meta- and reinforcement learning, these systems manage data selection as a \textit{self-evolving} data learning loop, dynamically curating the training stream for peak efficiency in large-scale model training.




\subsection{\textbf{Distributed Training Techniques}}
\label{subsec:distributed_training_techniques}

Distributed training leverages multiple workers to maximize computational resources, accelerate training, and improve model accuracy. It addresses the computational and memory limitations of a single device but introduces three major challenges: (1) Single devices cannot meet the computational demands of modern large-scale training; (2) Model parameters exceed the memory limits of a single device; (3) Distributed training suffers from high communication costs due to frequent synchronization. These challenges are addressed through multiple \textit{parallelization} and \textit{mixed-precision training} strategies in distributed training.

\subsubsection{\textbf{Data and Model Parallelism}}
\label{subsubsec:data_and_model_parallelism}


\noindent\textbf{\textit{Data Parallelism}}. As a fundamental technique for improving training throughput, data parallelism~\cite{li2020pytorch}  replicates the model across all workers, with each processing a distinct subset of the dataset. To maintain weight consistency, workers periodically synchronize their gradients. ZeRO~\cite{rajbhandari2020zero} (Zero Redundancy Optimizer), introduced by the Deep-Speed~\cite{rasley2020deepspeed}, significantly reduces memory usage  by optimizer states partitioning, gradients partitioning, and model parameters partitioning, eliminating redundant storage across devices. However, its main limitation lies in the substantial communication overhead that arises with extremely large models.

\noindent\textbf{\textit{Model Parallelism}}. When a model exceeds the memory capacity of a single device, model parallelism becomes necessary. Two prominent strategies are Tensor Parallelism (TP)~\cite{shoeybi2019megatron} and Pipeline Parallelism (PP)~\cite{huang2019gpipe}. TP decomposes weight matrices within specific layers (\eg, attention or feed-forward networks) across devices, enabling parallel computation. However, each forward and backward pass necessitates full all-reduce operations to synchronize partial results, introducing notable communication overhead. In contrast, PP distributes entire layers across devices arranged in a sequential pipeline, reducing per-device memory load at the expense of potential pipeline bubbles. While both approaches mitigate memory constraints by distributing parameters and computation, they incur additional communication costs.

\noindent\textbf{\textit{Hybrid Parallelism}}. Hybrid parallelism integrates data, tensor, and pipeline parallelism to efficiently train ultra-large-scale models, such as those with hundreds of billions of parameters, as exemplified by 3D parallelism~\cite{narayanan2021efficient}.

\subsubsection{\textbf{Mixed-Precision Training}}  
\label{subsubsec:mixed_precision_training}


During large-scale model training, GPU memory consumption primarily arises from four components: model parameters, optimizer states, intermediate activations, and temporary buffers. To address this constraint, mixed precision training has evolved from an optional optimization to a pivotal strategy for large-scale model development. It effectively overcomes critical bottlenecks in GPU memory and computational throughput by utilizing lower-precision formats for most operations (\eg, matrix multiplications) to reduce memory usage and accelerate computation. 

To preserve numerical stability, a master copy of the weights is maintained in a 32-bit floating point (FP32) for gradient accumulation and parameter updates. Additionally, loss scaling is applied: the loss is multiplied by a scaling factor before backpropagation to prevent small gradients from vanishing in FP16, and these gradients are unscaled before updating the master weights. In practice, studies such as~\cite{micikevicius2018mixed} have adopted FP16 or Brain Floating Point (BF16)~\cite{workshop2022bloom} formats, significantly decreasing memory usage while maintaining training performance.


\vspace{-8pt}
\begin{figure}[h!]
\centering
\includegraphics[width=0.48\textwidth]{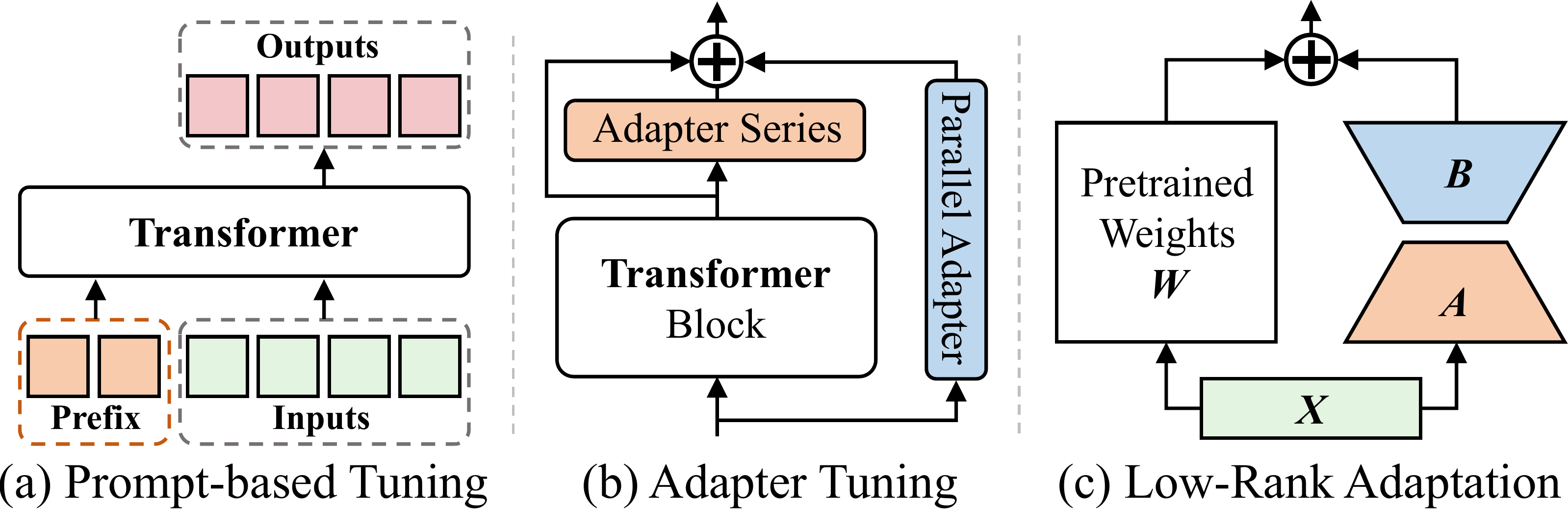}
\vspace{-5pt}
\caption{Illustration of the model architectures of three different parameter-efficient fine-tuning methods: (a) Prompt-based Tuning, (b) Adapter Tuning, and (c) Low-Rank Adaptation.}
\label{PEFT}
\vspace{-2mm}
\end{figure}

\subsection{\textbf{Parameter Efficient Fine-Tuning}}
\label{subsec:parameter_efficient_fine_tuning}

Parameter-Efficient Fine-Tuning (PEFT) refers to techniques that adapt large pretrained models to downstream tasks by updating only a small fraction of parameters or by inserting lightweight trainable modules while keeping most of the backbone frozen. The goals are to reduce training compute and memory, minimize storage and transmission of task-specific weights, and enable modular multi-task deployment. PEFT methods are widely used in language, vision, and multimodal models; they differ in where trainable capacity is placed (input-level prompts~\cite{brown2020language, lester2021power, li2021prefix}, modular layer inserts~\cite{houlsby2019parameter,hetowards}, or low-rank parameter subspaces~\cite{hu2022lora, wei2024flexora}), in trade-offs between parameter budget and final accuracy, and in deployment convenience. 
Below we summarize three common PEFT families: \textit{Prompt-based Tuning}, \textit{Adapter Tuning}, and \textit{Low-Rank Adaptation}, and provide representative references for each PEFT method.


\subsubsection{\textbf{Prompt-based Tuning}}
\label{subsubsec:prompt-based_tuning}

Prompt-based Tuning exploits the fact that pretrained language models can often be steered toward desired downstream behavior by suitable prompts. 
Early work with hand-crafted discrete prompts demonstrated strong few-shot capabilities in large language models~\cite{brown2020language}. 
To make prompts trainable and more general, continuous (soft) prompt methods were proposed. Specifically, 
Prompt Tuning~\cite{lester2021power} prepends a small set of learnable embeddings to the model input and trains only those embeddings, as shown in Figure~\ref{PEFT}(a). 
Prefix Tuning~\cite{li2021prefix} injects learnable embeddings into each transformer layer's attention mechanism so the prefix can influence internal computations without modifying model weights. 
P‑tuning and P‑tuning v2~\cite{liu2024gpt, liu2022p} introduce richer parameterizations and optimization strategies for continuous prompts and show effectiveness across generation and classification tasks. 
Prompt-based methods are attractive due to their minimal parameter footprint and easy per-task storage, but their effectiveness can be sensitive to model scale, prompt length, and initialization; they may underperform other PEFT methods on smaller backbones or on tasks requiring deeper representational changes. Recent work addresses multi-task prompt learning, initialization heuristics, and optimization improvements to boost stability and generalization.

\subsubsection{\textbf{Adapter Tuning}}
\label{subsubsec:adapter_tuning}


As shown in Figure~\ref{PEFT}(b), Adapter Tuning inserts compact, trainable adapters into frozen layers and trains only those modules. 
Typical adapters use a bottleneck architecture (down-projection → nonlinearity → up-projection), adding a few parameters while enabling flexible transformations of layer activations. 
The Series Adapter~\cite{houlsby2019parameter, fu2021learn}, which places adapter blocks sequentially inside each Transformer block, was an early and widely used design. 
To reduce latency and preserve parallelism, Parallel Adapters~\cite{hetowards} place a lightweight side branch that runs in parallel with the original sublayer and fuse its output with the main branch; these designs reduce inference overhead while retaining much of serial adapters' performance. 
Extensions such as AdapterFusion~\cite{pfeiffer2021adapterfusion} and related methods~\cite{wang2022adamix, he2023mera} fuse multiple task adapters for transfer. Current directions include automated adapter architecture search~\cite{lawton2023neural, zhou2024autopeft}, adapter compression~\cite{he2022sparseadapter}, and adapting adapter ideas to vision and multimodal transformers~\cite{zhang2023adding, advclip}.

\begin{figure*}[t!]
\centering
\includegraphics[width=1\textwidth]{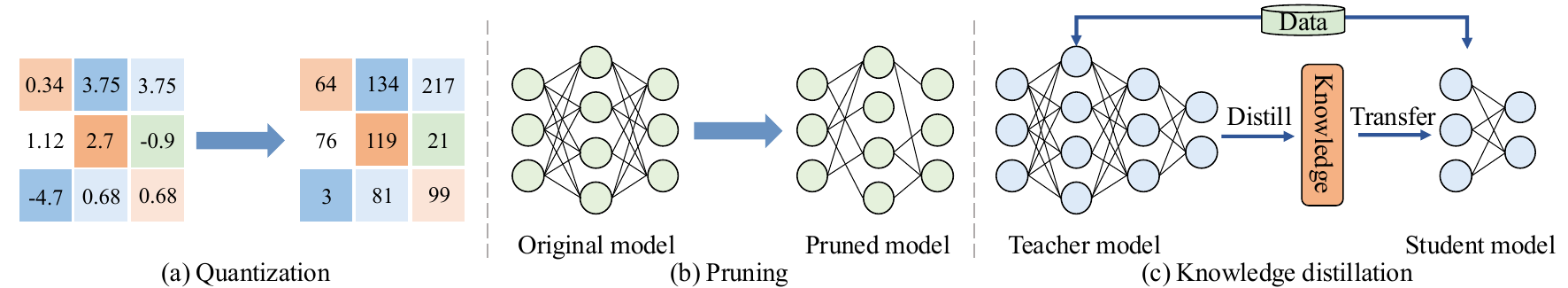}
\vspace{-18pt}
\caption{Illustration of techniques commonly employed for model compression. Mainly include (a) quantization, (b) pruning, and (c) knowledge distillation.}
\label{compression}
\vspace{-10pt}
\end{figure*}

\subsubsection{\textbf{Low-Rank Adaptation}}
\label{subsubsec:low_rank_adaptation}

Low-Rank Adaptation constrains the learned update to a weight matrix to lie in a low-dimensional subspace, expressing the update as a product of two small matrices. As shown in Figure~\ref{PEFT}(c), LoRA~\cite{hu2022lora} is the canonical example. Instead of updating a full weight matrix $\bm{W}$, LoRA freezes $\bm{W}$ and learns low-rank matrices $\bm{A}$ and $\bm{B}$ so that the effective weight becomes \(\bm{W} + \alpha \bm{A} \bm{B}\) during training (where \(\alpha\) is a scaling factor). 
When applied to a linear projection that maps an input \(\bm{X}\) via \(\bm{W}\), the LoRA update modifies this mapping to \(\bm{W}\bm{X} + \alpha \bm{A}\bm{B}\bm{X}\). 
LoRA is typically applied to critical linear projections (\eg, query/key/value or feed-forward projections) in Transformer blocks. The approach offers strong parameter–performance trade-offs for large models, requires little additional GPU memory, and can be merged into base weights for inference, simplifying deployment. Since its introduction, LoRA has inspired many variants: selective layer application~\cite{ zhou2025lora}, dynamic/adaptive rank selection~\cite{shinwari2025ard, lu2024adaptive}, and hybrid schemes combining LoRA with adapters or prompts~\cite{xiao2024hivg, chen2024llama}. 
LoRA's engineering advantages, including compatibility, easy merging into base weights, and straightforward deployment, have driven broad industrial adoption and spurred numerous benchmarks and open-source implementations such as Hugging Face tools~\cite{wolf-etal-2020-transformers} and the PEFT library~\cite{peft_tools}. Beyond these widely used libraries, several open-source ecosystems further simplify LoRA-style fine-tuning. 
FastChat~\cite{fastchat} provides a lightweight framework for instruction tuning and chat model serving with built-in LoRA adapter support. 
Unsloth~\cite{unsloth} offers optimized kernels and quantized LoRA training, reducing memory and compute demands. 
ColossalAI~\cite{Colossal} integrates LoRA with system-level optimizations such as parallelism and memory partitioning, enabling efficient fine-tuning of large models on limited hardware. 
These toolkits broaden practical PEFT adoption by lowering resource requirements and easing deployment.



\subsection{\textbf{Model and Computational Compression}}
\label{subsec:model_compression}

Model and computational compression aim to reduce the computational size by eliminating redundant information, thus enhancing both storage efficiency and computational performance during inference. As illustrated in Figure~\ref{compression} and Figure~\ref{speculative_decoding}, key techniques encompass \textit{low-precision quantization}, \textit{model pruning}, \textit{knowledge distillation}, and \textit{speculative decoding}. The following subsections will delve deeper into these techniques and their role in enhancing model performance during inference.

\subsubsection{\textbf{Low-Precision Quantization}}
\label{subsubsec:low_precision_quantization}

Low-precision quantization has become a crucial technique for improving the training and inference efficiency of large-scale deep learning models \cite{wang2025empowering}. Originally, 32-bit floating point (FP32) was the standard for model training, but the growing size of models and datasets has made its computational cost prohibitive \cite{yuan2025give}. To mitigate this issue, lower-precision formats such as FP16, BF16, and more recently FP8 and INT8 have been widely adopted. These formats offer benefits including reduced memory consumption and faster computation \cite{van2023fp8}. Nevertheless, numerical overflow and precision degradation remain critical challenges, especially for long-sequence training and large-scale inference tasks \cite{peng2023fp8}.

As quantization techniques continue to advance, several methods have been proposed to improve stability and performance. LLM.int8()~\cite{dettmers2022gpt3} demonstrates that INT8 quantization can significantly reduce memory usage while preserving near-FP16 accuracy, and LLM-QAT~\cite{liu2023llm} incorporates quantization into training to enhance robustness to low precision. To further address instability, dynamic precision strategies such as Fallback Quantization \cite{zhang2025accurate} temporarily elevate precision to prevent overflow, while ShiftQuant \cite{guo2024towards} enables sub-8-bit training via gradient estimation and L1 normalization. More recently, the pursuit of extreme efficiency has spurred interest in INT4 and even 1-bit quantization. Frameworks such as OneBit \cite{xu2024onebit} and BitNet \cite{wang2023bitnet} illustrate the feasibility of binarizing weights with minimal accuracy loss, and techniques like ParetoQ \cite{liuparetoq} and ABQ-LLM \cite{zeng2025abq} explore arbitrary-bit configurations to flexibly balance precision and efficiency.

\subsubsection{\textbf{Model Pruning}}
\label{subsubsec:model_pruning}


Model pruning serves as a pivotal optimization strategy in large language models, enabling significant enhancements in inference speed by selectively removing less critical parameters while preserving core functionality \cite{liu2025pruning}. This approach reduces the computational footprint during deployment, allowing for faster processing on resource-constrained devices without necessitating extensive retraining \cite{hou2025instruction}.

Recent surveys on efficient LLMs highlight pruning innovations addressing inference bottlenecks like high FLOPs and memory demands \cite{whitmore2025efficient,liu2025pat}. One-shot methods like SparseGPT \cite{frantar2023sparsegpt} induce up to 50\% sparsity in models such as OPT-175B, reducing inference time via parameter elimination without iterative fine-tuning, aiding latency-sensitive deployments. Activation-aware techniques like Wanda \cite{sun2023simple} prune based on weight-activation interactions, yielding $1.24\times$ speedups by cutting redundant computations for dynamic querying. Structured approaches such as SliceGPT \cite{ashkboos2024slicegpt} use PCA to remove low-importance components, achieving $1.87\times$ acceleration and 30\% compression in Transformers, boosting scalability for mobile and distributed systems. Semi-structured patterns in E-Sparse \cite{li2023sparse} align with hardware like Tensor Cores for $1.53\times$ throughput gains, linking sparsity to practical utilization. These advances mitigate LLM inefficiencies, enabling broader use in resource-constrained environments.

\subsubsection{\textbf{Knowledge Distillation}}
\label{subsubsec:knowledge_distillation}

Knowledge distillation emerges as a sophisticated refinement technique for large language models, facilitating accelerated inference by transferring distilled insights from a robust teacher model to a more compact student counterpart, thereby slashing computational demands during runtime \cite{yang2024survey,xu2024survey}. This method optimizes deployment efficiency, enabling quicker response times on hardware with limited capabilities without the need for exhaustive reconfiguration \cite{takrouri2025knowledge}.

Contemporary research highlights black-box and white-box strategies for boosting LLM inference velocity. For example, Orca 2 \cite{mitra2023orca} uses synthetic data from teacher models to enhance reasoning in smaller students, achieving $2–3\times$ gains over Vicuna on complex tasks with minimal data and 40\% acceleration in reasoning pipelines. TinyLLM \cite{kandala2024tinyllm} aggregates multi-teacher knowledge into a small student, delivering up to 30\% inference speedup on edge devices while preserving 90\% accuracy across benchmarks. DA-KD \cite{citation-0} applies difficulty-aware sampling to focus on challenging samples, yielding $1.8\times$ throughput improvements for efficient LLMs in real-time applications like chatbots. Symbolic Chain-of-Thought Distillation refines symbolic reasoning paths, allowing small models to emulate step-by-step thinking with 25\% reduced latency in logical scenarios \cite{li2023symbolic}.

\subsubsection{\textbf{Speculative Decoding}}
\label{subsubsec:speculative_decoding}

\begin{figure}[h!]
\centering
\includegraphics[width=0.43\textwidth]{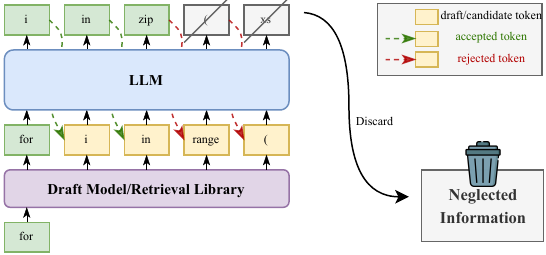}
\vspace{-8pt}
\caption{Illustration of the speculative decoding mechanism \cite{luo2024turning}. The speculative decoding methods draft some tokens and verify them in parallel in one decoding step.}
\vspace{-3mm}
\label{speculative_decoding}
\end{figure}

Speculative decoding functions as an innovative speedup tactic for large language models, hastening inference by employing a lightweight drafter to propose token sequences that a primary verifier then assesses in parallel, thereby curtailing sequential computations and elevating generation rates, as shown in Figure~\ref{speculative_decoding}. 

\vspace{-5pt}

Contemporary explorations in speculative decoding for LLMs focus on draft-verification synergies to enhance throughput in resource-intensive inference. For example, Confidence-Modulated Speculative Decoding \cite{sen2025confidence} applies adaptive confidence thresholds for dynamic drafting lengths, yielding up to $2.5\times$ faster decoding in variable-context LLMs with noise robustness. Recurrent Drafter \cite{cheng2024recurrent} uses RNN-based drafters for sequential efficiency, achieving $3\times$ speedups in long-form generation by cutting verification overheads in Llama variants. Efficient Multi-sample Speculative Decoding \cite{ni2024ems} leverages parallel sampling to raise acceptance rates, providing $2.2\times$ inference gains and reduced latency in NAACL-benchmarked LLMs. SpecEE \cite{xu2025specee} merges speculative execution with early exiting, cutting computational demands by 4\% and boosting on-device throughput for edge-deployed LLMs.

\vspace{7pt}
\section{Green Computing AI Chips and Hardware-Software Co-Design}
\label{sec:chapter4}

As Large Language Models (LLMs) continue to scale in complexity and deployment, the demand for high-performance and efficient hardware has intensified. Traditional acceleration platforms, while effective, have exposed growing concerns over energy consumption, carbon emissions, and deployment scalability. In response, both academia and industry have invested heavily in the development of green computing solutions—from architectural innovations in AI chips to sustainable data center infrastructure \cite{budennyy2022eco2ai,strubell2020energy}. This chapter explores the evolving landscape of AI hardware, emphasizing mainstream chip ecosystems and emerging paradigms in energy-efficient hardware-software co-design.

\begin{figure}[h!]
\centering
\includegraphics[width=0.31\textwidth]{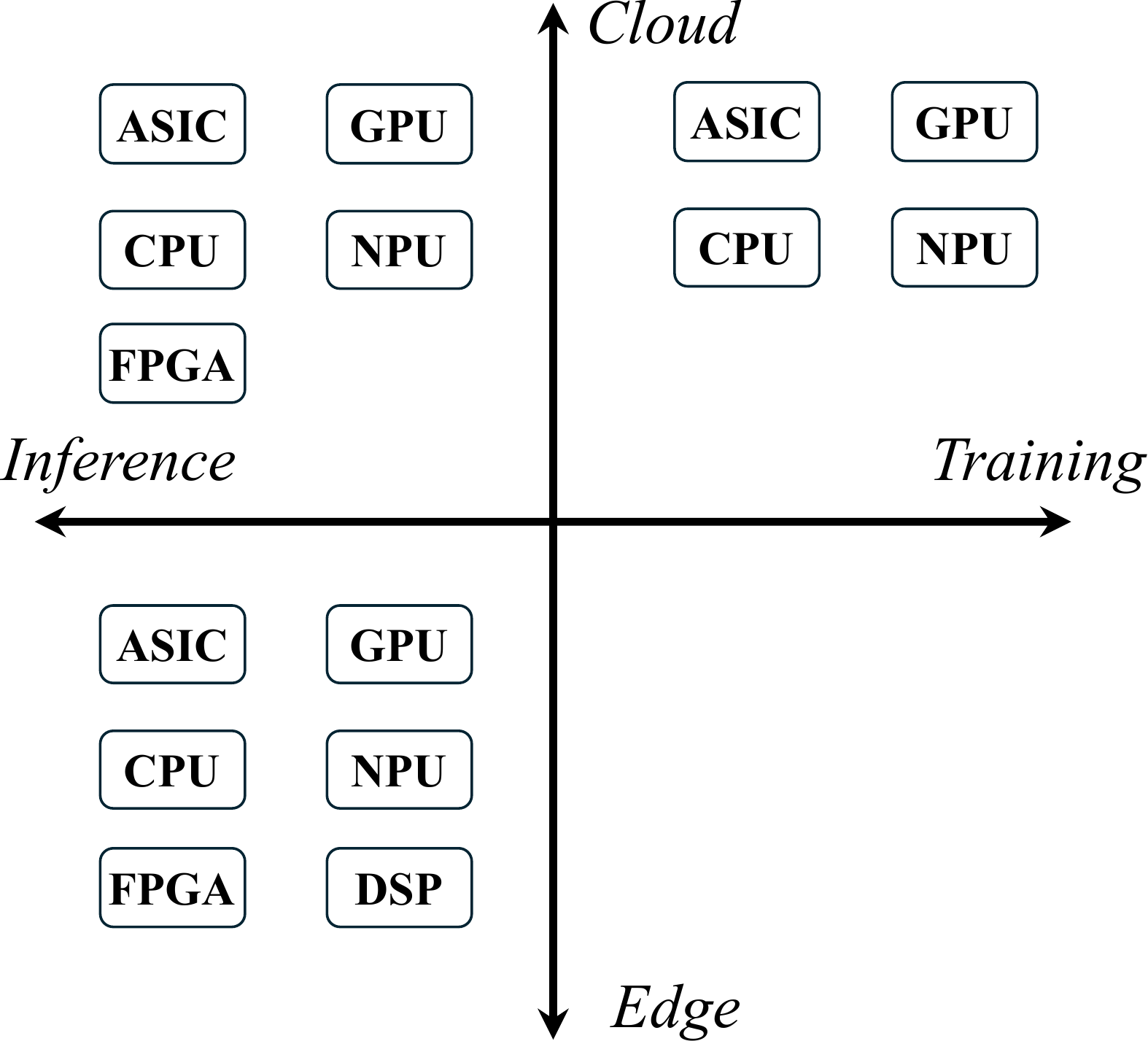}
\vspace{-5pt}
\caption{Application Categories of AI Chips.}
\label{fig:chap4_ai_chip_type}
\vspace{-12pt}
\end{figure}

\subsection{Mainstream AI Chips and Application Ecology}
\label{subsec:mainstream_ai_chips_and_application_ecology}

\subsubsection{\textbf{Classification and Comparison of AI Chips}}
\label{subsubsec:classification_and_comparison_of_ai_chips}

As shown in Figure~\ref{fig:chap4_ai_chip_type}, AI chips can be broadly categorized based on their deployment environments (\ie, cloud or edge), and their computational roles in training or inference. Central Processing Units (CPUs), while flexible, are generally inefficient for large-scale matrix operations. Graphics Processing Units (GPUs), such as NVIDIA's A100 and H100, have become the de facto standard for training LLMs due to their massive parallelism and rich software ecosystem (\eg, CUDA). Tensor Processing Units (TPUs)~\cite{jouppi2017datacenter}, developed by Google, are custom ASICs optimized for matrix multiplication and serve as a backbone for models like GPT and PaLM. Field-Programmable Gate Arrays (FPGAs) offer reconfigurability and are well-suited for edge inference but require significant design effort \cite{chen2016eyeriss}. Digital Signal Processors (DSPs) and Application-Specific Integrated Circuits (ASICs) provide optimized performance for specific workloads, especially in low-power scenarios \cite{hu2022survey}.

\begin{table}[t!]
\setlength\tabcolsep{4pt}
\caption{Global Representative AI Chip Enterprises and Its Product.}
\label{tab:chap4_chips}
\resizebox{0.88\linewidth}{!}{
\begin{tabular}{c|c|c|c|c}
\hline
Company & Typical Chips   & Year &  Arch  & Function        \\ \hline
\multicolumn{5}{c}{International AI chip enterprises}       \\ \hline
 NVIDIA    & H100         & 2022 & GPU  & cloud training    \\ 
 AMD       & MI300        & 2023 & GPU  & cloud training    \\ 
 Intel     & Gaudi 3      & 2024 & NPU  & cloud training    \\ 
 Google    & TPUv5        & 2023 & ASIC & cloud training    \\ 
 Qualcomm  & Cloud AI100  & 2020 & ASIC & cloud inference   \\ 
 Apple     & M3           & 2023 & ARM  & edge inference    \\ 
 Samsung   & Exynos 2100  & 2021 & ARM  & edge inference    \\
 IBM       & TrueNorth    & 2015 & -- &  edge inference     \\ \hline 
\multicolumn{5}{c}{Domestic AI chip enterprises}            \\ \hline
 Cambricon & MLU 590      & 2024 & ASIC & cloud training    \\ 
 HUAWEI    & Ascend 910   & 2019 & NPU  & cloud training    \\ 
 Horizon   & Journey 6    & 2023 & ARM  & edge inference    \\ 
 T-head    & Hanguang 800 & 2019 & FPGA & cloud inference   \\ 
 Baidu     & Kunlun 2     & 2022 & FPGA & cloud inference   \\ \hline
\end{tabular}}
\centering
\vspace{-10pt}
\end{table}

\subsubsection{\textbf{Global AI Chip Ecosystems}}
\label{subsubsec:global_ai_chip_ecosystems}


As shown in \cref{tab:chap4_chips}, leading international companies such as NVIDIA, Google, AMD, and Intel dominate the AI hardware market. Firstly, NVIDIA's GPU platform combines rapidly evolving silicon architectures, such as Ampere, Hopper, and Blackwell, with a vertically integrated software stack, including CUDA, CUTLASS, cuBLAS/cuDNN, NCCL, and Triton Inference Server, and system-level interconnects like NVLink and NVSwitch, enabling scalable training across tens of thousands of accelerators. Secondly, Google’s TPU series~\cite{jouppi2017datacenter}, from TPU v2/v3 to v5, demonstrates the datacenter efficiency of AI-specific ASICs by leveraging systolic matrix engines, wafer-scale torus interconnects, and optical circuit switches for dynamic topology reconfiguration. These innovations achieve high model FLOPs utilization in large-scale training while minimizing energy consumption per operation. Google integrates TPUs into its cloud infrastructure, supporting the training of cutting-edge models such as Gemini \cite{team2023gemini}. Meanwhile, AMD's MI300 series and Intel's Gaudi 3 accelerators are broadening the competitive landscape by emphasizing mixed-precision computing and energy-aware scheduling. These platforms are supported by comprehensive toolchains and SDKs that enable developers to optimize LLM workloads across both training and inference stages \cite{wu2025development}.


\subsubsection{\textbf{Domestic AI Chip Development and Ecosystems}}
\label{subsubsec:domestic_ai_chip_development_ecosystems}


As shown in \cref{tab:chap4_chips}, China’s AI chip ecosystem has diversified across cloud and edge scenarios \cite{wu2025development}. Cloud- and edge-oriented NPUs from Cambricon \cite{liu2016cambricon} (\eg, MLU series) target training and inference with compiler toolchains and operator libraries integrated into domestic AI frameworks. Horizon Robotics focuses on edge autonomy with high-efficiency SoCs (Journey/Sunrise), combining perception and planning accelerators for automotive ADAS/AD. Huawei’s Ascend (\eg, 910/310) provides a vertically integrated stack for cloud and edge with the CANN/AscendCL toolchain, MindSpore framework, and model zoo integration; the series have been deployed for general AI services and domain-specific inference. In special-purpose domains, Bitmain’s experience in ultra–high-efficiency matrix engines informs inference ASIC design practices, while T-Head (Alibaba) advances RISC‑V based CPUs (XuanTie) and cloud inference chips (Hanguang) within a broader open-source and cloud ecosystem. These vendors are increasingly developing end-to-end stacks, including compilers, graph optimization, quantization, and runtimes, as software maturity has become a decisive factor in achieving effective energy efficiency at the application level

\subsubsection{\textbf{Energy and Performance Perspectives}}
\label{subsubsec:energy_and_performance_perspectives}

The energy efficiency of AI chips is an increasingly critical metric, often measured in TOPS/W (Tera-operations per Second per Watt). For instance, while NVIDIA’s H100 achieves state-of-the-art throughput, its power envelope exceeds 700W, posing challenges for widespread deployment. In contrast, edge ASICs like those from Bitmain or Apple’s Neural Engine offer much higher energy efficiency for inference. The software stack plays a pivotal role in green computing. Frameworks like TensorRT, TVM, and MindSpore enable operator-level optimizations, while sparsity-aware libraries reduce redundant computation. Matching model architecture (\eg, Transformer variants) to chip capabilities is essential for maximizing energy-performance trade-offs.

\subsection{\textbf{Inference memory optimization}}
\label{subsec:inference_memory_optimization}

Inference memory optimization serves as a critical advancement in large language models, achieved through software-hardware co-design that accelerates inference by dynamically reallocating storage resources via intelligent caching mechanisms and offloading inactive data to secondary storage layers, thereby reducing peak memory demands without sacrificing throughput. Key strategies encompass KV cache optimization, memory offloading, and early exiting.

\begin{figure}[t!]
\centering
\vspace{-6pt}
\includegraphics[width=0.42\textwidth]{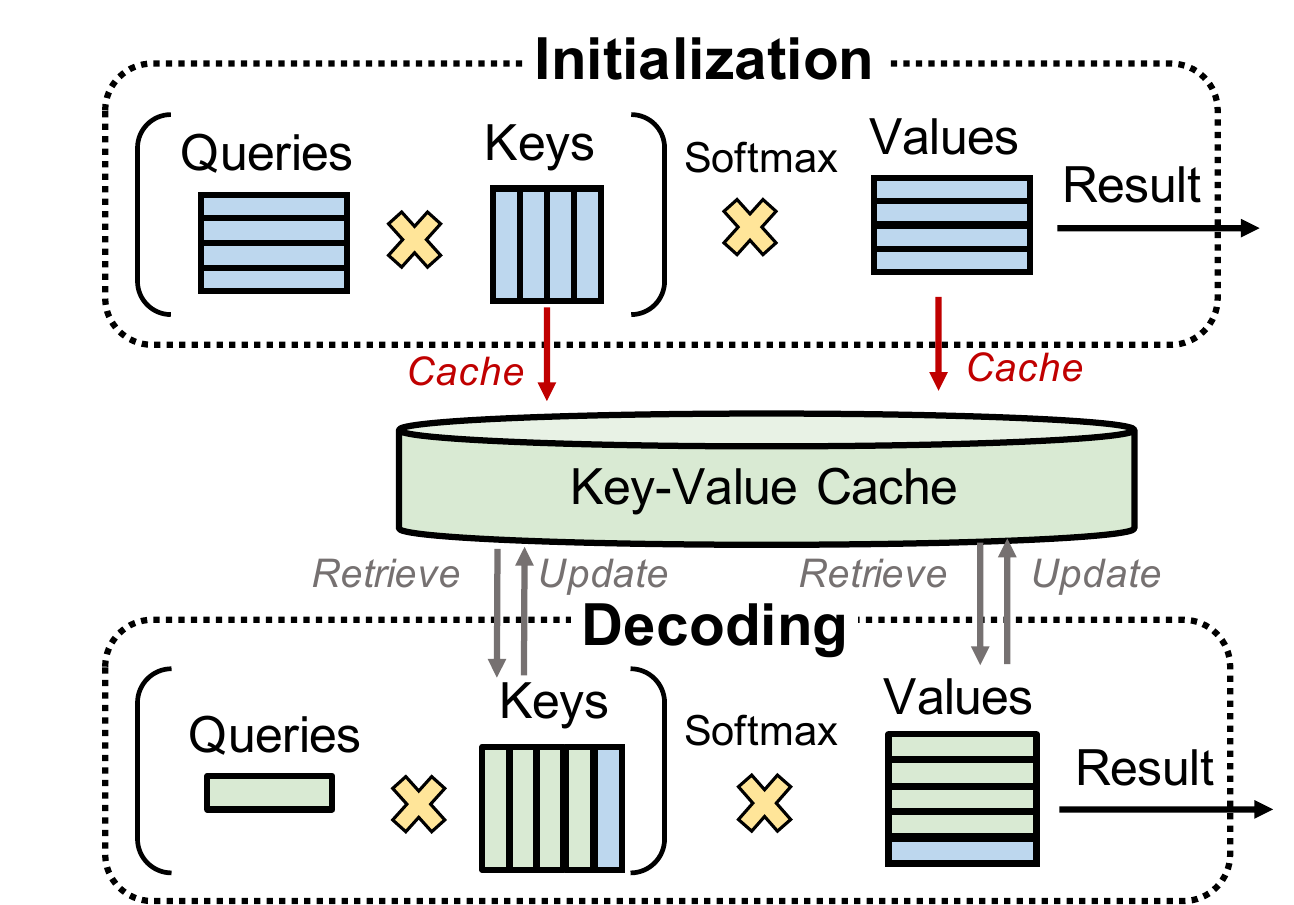}
\vspace{-4pt}
\caption{Illustration of the key-value caching mechanism \cite{wu2023fast}.}
\vspace{-9pt}
\label{KV}
\end{figure}

\subsubsection{\textbf{KV Cache Optimization}}
\label{subsubsec:kv_cache_optimization}


KV cache optimization emerges as a pivotal acceleration technique in large language models, realized through software-hardware co-design that boosts inference efficiency by compressing key-value stores using selective retention and precision reduction methods, thereby mitigating memory bottlenecks in long sequences while preserving accuracy \cite{liu2025kv,li2024survey}. Fig.~\ref{KV} demonstrates the key-value cache usage in both phases. During the initialization phase, the LLM generates the key-value cache for each token in the input prompt. In the subsequent decoding phase, the LLM only needs to compute the query, key, and value of one newly generated token, leveraging the precomputed key-value cache to facilitate the process step by step.

Research on LLM-accelerated KV cache optimization categorizes strategies into tag-level eviction and model-level quantization, using hardware-aware designs to reduce memory overhead in long-context scenarios. For instance, CacheGen \cite{liu2024cachegen} applies custom tensor encoding via distributional properties, enabling 2-4$\times$ faster context loading and streaming for Llama-2-70B in high-throughput serving. LaCache \cite{shi2025lacache} uses ladder-shaped caching with training-free hierarchical retention, boosting inference speed by 1.5-2$\times$ and reducing cache size by 40\% for Mistral-7B in long-sequence generation. FastGen \cite{ge2023model} employs LLM profiling for adaptive compression across attention heads, halving memory usage without quality loss and accelerating decoding by 30\% in dynamic workloads. Gao \textit{et al.} propose hybrid sparsity-quantization frameworks, achieving up to $3\times$ throughput gains by minimizing recomputation in extended contexts \cite{gao2025rethinking}.


\subsubsection{\textbf{Memory Offloading}}
\label{subsubsec:memory_offloading}

Memory offloading emerges as a pivotal technique for accelerating LLM inference on resource-constrained hardware, realized through software-hardware co-design that enables the execution of massive models by dynamically swapping parameters between high-speed GPU memory and slower but more capacious external storage such as CPU DRAM or NVMe SSDs \cite{davies2025efficient}. 

Recent advancements refine offloading strategies to balance bandwidth and compute demands. For example, Aqua \cite{vijaya2025aqua} employs network acceleration in multi-GPU clusters to cut paging overheads in inference state transfers, achieving up to $2\times$ speedups via RDMA weight prefetching. HeadInfer \cite{luo2025headinfer} uses head-wise KV cache offloading, partitioning attention heads across devices to reduce full-layer storage and increase memory efficiency by 40\% without accuracy loss. For MoE architectures, a latency-hiding scheme \cite{wang2025accelerating} overlaps expert activations with data transfers, minimizing idle times for seamless scaling on heterogeneous setups. SpecOffload \cite{zhuge2025specoffload} applies speculative partial offloading to tap underused GPU capacity, predicting and caching low-activation parameters on CPU for $1.5$-$3\times$ faster iterative decoding. Complementary methods, such as heterogeneous speculative decoding, multilevel pretraining offloads, and dynamic token pruning, illustrate the growing inference optimization ecosystem.


\subsubsection{\textbf{Early Exiting for Inference Acceleration}}
\label{subsubsec:early_exiting}

Early exiting has become a cornerstone of conditional computation in LLM inference acceleration, dynamically halting autoregressive decoding at intermediate layers when token predictions meet predefined confidence criteria, thus curbing computational overhead and enabling real-time applications on diverse hardware without compromising semantic integrity \cite{miao2024efficient}.

Over the past few years, innovations like AdaInfer \cite{fan2024not} and FREE \cite{bajpai2025free} have paved the way for more adaptive mechanisms, while broader efforts in speculative integration enhance throughput. SpecEE \cite{xu2025specee} deploys a speculation-driven engine that anticipates exit points via auxiliary predictors, verifying drafts in parallel to yield up to $2.5\times$ faster generation on GPUs by minimizing redundant layer traversals. Finally, SPADE introduces a hybrid algorithm fusing confidence monitoring with space-aligned decoding, projecting intermediate states to align with deeper representations and enabling seamless exits that preserve coherence, boosting efficiency by $2\times$ in edge deployments \cite{zheng2025hybrid}.

\subsection{\textbf{Cross-platform deployment and adaptation}}
\label{subsec:cross_platform_deployment_and_adaptation}

Cross-platform deployment and adaptation for large-scale models facilitates efficient execution across heterogeneous hardware like CPUs, GPUs, and edge devices by addressing compatibility and optimization challenges through three core strategies: unified inference frameworks, automatic operator scheduling, and key operator discretization. The following subsections explore these techniques and their impacts on model performance.

\subsubsection{\textbf{Unified Inference Framework}}
\label{subsubsec:unified_inference_framework}


Unified inference frameworks streamline LLM cross-platform deployment by abstracting hardware details into a cohesive API, enabling seamless transitions across CPUs, GPUs, and edge devices while maintaining performance.

Key approaches include LLMBox \cite{tang2024llmbox}, which unifies training, inference, and evaluation pipelines with modular extensions for backends like PyTorch and TensorRT to reduce deployment overheads. HERMES \cite{bambhaniya2025understanding} supports multi-stage pipelines with heterogeneous clients, dynamically batching requests across concurrent models to optimize distributed resource use. ScaleLLM \cite{yao2024scalellm} emphasizes end-to-end efficiency via operator fusion and paged attention for scalable serving beyond standard inference. Together, these reduce porting efforts by up to 50\% in multi-tenant settings. Edge latency-aware LLM customization \cite{tian2025clone} uses distillation in unified backends for privacy-preserving inference on constrained hardware. Overall, these frameworks boost adaptability, yielding $2$-$3\times$ throughput improvements across ARM and x86 ecosystems.

\subsubsection{\textbf{Automatic Operator Scheduling}}
\label{subsubsec:automatic_operator_scheduling}

Automatic operator scheduling optimizes LLM inference by dynamically allocating computational kernels to heterogeneous hardware, mitigating bottlenecks in cross-platform execution through runtime profiling and predictive mapping. 

The Past-Future Scheduler in LightLLM \cite{gong2025past} balances historical and speculative workloads under SLA constraints, prioritizing token sequences to improve goodput by $1.5\times$ in shared clusters. Decentralized serving task scheduling \cite{wu2025task} applies heuristic algorithms to distribute inference across edge nodes, enabling fault-tolerant adaptation for large-scale deployments. DynamoLLM \cite{stojkovic2025dynamollm} uses reconfiguration loops to optimize cluster topologies for energy efficiency, aligning operator dispatch with SLOs through reinforcement learning policies. These approaches handle multi-tenancy by integrating scheduling with KV cache management.

\subsubsection{\textbf{Key Operator Discretization}}
\label{subsubsec:key_operator_discretization}


Key operator discretization enhances cross-platform LLM adaptation by decomposing complex kernels into modular, hardware-agnostic units, allowing fine-grained quantization and fusion for efficient porting across accelerators. 

ClusterFusion \cite{kurniawan2023clusterfusion} scales fusion primitives to cluster-level operators, discretizing attention and linear layers for distributed inference with $2\times$ memory savings. FlashDecoding++ \cite{hong2024flashdecoding++} uses asynchronous kernel fusion in LLM decoding, discretizing GEMM operations for flat optimization and heuristic prefetching. Qtile \cite{zhang2025qfactory} accelerates quantized serving through tile-based operator discretization, integrating low-bit formats with fusion to minimize overhead in multi-precision setups and mitigate variances on ARM and NVIDIA hardware. KPerfIR \cite{guan2025kperfir} advances a compiler-centric ecosystem for GPU kernel fusion, discretizing operators via 4D parallelism to balance workloads in LLM pipelines. These methods emphasize hardware-agnostic optimizations, enabling broader LLM accessibility in diverse computational environments.

\subsection{Energy-Efficient Hardware System}
\label{subsec:energy_efficient_hardware_system}



Beyond AI chips, some hardware directions are seeking step-function improvements in performance per watt by reducing energy consumption associated with data movement and cooling, and by tailoring computation to model structure.

\subsubsection{\textbf{Emerging Green Chip Technologies}}
\label{subsubsec:emerging_green_chip_technologies}

To overcome the limitations of the Von Neumann architecture, where data shuttling between memory and compute units incurs substantial energy overhead, In-Memory Computing (IMC) has emerged as a promising alternative \cite{wan2022compute}. IMC performs computation within the memory array itself, significantly reducing data movement. Architectures like TranCIM, RIME, and SmartInfinity showcase how compute-in-memory enables energy-efficient matrix operations, particularly for low-bit quantized models. Additionally, neuromorphic chips \cite{basu2022spiking} such as IBM’s TrueNorth and Intel’s Loihi simulate brain-like spiking neuron behavior \cite{davies2021lessons}, enabling ultra-low-power inference. Optical computing and quantum accelerators also show promise, though they remain in experimental stages \cite{shastri2021photonics}. These paradigm-shifting technologies aim to support future LLMs with orders-of-magnitude improvements in energy efficiency.

\subsubsection{\textbf{Decentralized Green Learning}}
\label{subsubsec:decentralized_green_learning}

Decentralized learning paradigms further reduce the environmental cost of AI by curbing data movement and right‑sizing compute. Federated Learning (FL) offers a distributed training paradigm where models are trained locally on edge devices and only parameter updates are exchanged. This approach not only enhances data privacy but also minimizes the need for large-scale data transmission to central servers, reducing overall energy consumption \cite{sun2021decentralized}. Recent studies show that federated fine-tuning of LMs can lead to competitive performance with significantly lower compute and communication costs, especially when combined with techniques like LoRA and quantization-aware training. Besides, collaborative inference systems like PETALS \cite{borzunov2023petals} propose decentralized model hosting, where users share compute workloads, enhancing both accessibility and sustainability.


\subsubsection{\textbf{Green Data Centers}}
\label{subsubsec:green_data_centers}

Data centers hosting LLM workloads must address their rapidly growing carbon footprints \cite{cao2023data}. One strategy is geographical optimization, whereby data centers are located near renewable energy sources—such as hydropower in Sichuan, geothermal in Iceland, or solar farms in the American Southwest \cite{radovanovic2022carbon}. China's “East Data, West Computing” initiative exemplifies this approach, relocating compute-intensive LLM training to western regions with abundant green energy \cite{zhang2025decarbonizing}. Moreover, natural cooling strategies, including the use of ambient air or seawater for thermal management, are replacing traditional air-conditioning systems. Innovations in liquid immersion cooling and heat reuse (\eg, district heating) further contribute to reducing the energy overhead of data center operations.

\vspace{7pt}
\section{AI for Sustainability Applications}
\label{sec:chapter5}

As large models continue to evolve in scale and complexity, their potential to contribute to global sustainability efforts is becoming increasingly evident. Beyond concerns over their own carbon footprint, large models and AI systems are now being actively deployed to empower sustainable practices across multiple domains. This chapter explores recent advancements in AI applications for sustainability, focusing on efficient model architectures, high-throughput remote sensing, national-scale computing infrastructures, and domain-specific green innovations.



\subsection{Parallel Training and RL-Driven Reasoning Paradigm}
\label{subsec:deeepseek}

DeepSeek \cite{guo2025deepseek} exemplifies a dual focus on efficient large-model training and inference. In training, it leverages multi-level parallelism, including tensor, pipeline, and data parallelism, coupled with latency-hiding kernels, operator fusion, and topology-aware communication. These techniques maximize GPU utilization and minimize memory and communication overhead, aligning with best practices for distributed LLM training \cite{duan2024efficient}. On the inference side, DeepSeek adopts reinforcement learning at scale to improve reasoning efficiency. By learning structured decision-making policies and task-specific tool usage, the model reduces redundant computation and improves output quality per token and per joule. This approach reflects a broader trend: sustainable AI increasingly enhances energy efficiency through co-optimized training pipelines and inference-time policy learning, rather than relying solely on raw compute scaling \cite{oyewole2025sustainable}.

\subsection{Remote Sensing Interpretation: RS‑vHeat and ``Aerospace·Lingmou'' 3.0}
\label{subsec:remote_sensing_interpretation}


Remote sensing workloads require high-throughput, energy-efficient inference across multi-sensor platforms under stringent latency and power constraints. RS‑vHeat~\cite{hu2024rs} addresses these challenges through a physics-inspired architecture that models semantic propagation as heat diffusion~\cite{wang2025building}. Developed around the “Aerospace·Lingmou” 3.0 kernel, it replaces global attention with localized heat conduction operators, significantly reducing memory traffic while maintaining global receptive fields. This physical inductive bias enhances locality, facilitates aggressive kernel fusion and cache reuse, and achieves substantial efficiency gains: 84\% memory reduction, 24\% lower FLOPs, and 2.7$\times$ higher throughput compared to attention-based models, while maintaining state-of-the-art performance across optical, SAR, thermal, and hyperspectral modalities. The compact design enables efficient edge deployment and federated learning, reducing cloud dependency and emissions.

\subsection{China Computing NET (C$^2$NET): A National-Scale Sustainable Infrastructure}
\label{subsec:china_computing_net}

C$^2$NET envisions a unified, sovereign “compute network” that interconnects heterogeneous compute (GPU, NPU, CPU, FPGA) via high-speed links and programmable scheduling, providing on‑demand, policy‑aware capacity to national strategic workloads. As the digital economy’s substrate, C$^2$NET targets holistic sustainability in three directions. First, siting and energy: it supports “East Data, West Computing”, placing resource‑hungry clusters near renewable resources and abundant land while serving coastal demand over high‑performance backbones, cutting carbon intensity and land use \cite{zhang2025decarbonizing}. Second, orchestration: traffic‑ and topology‑aware schedulers allocate tasks to the cleanest, closest, and least‑congested sites, maximizing MFU and minimizing network energy per job~\cite{duan2024efficient}. Third, openness and resilience: a standardized interface for resource discovery, carbon-aware scheduling, and privacy‑preserving execution (\eg, federated analytics) lowers barriers to sharing clean capacity nationwide. As a result, C$^2$NET not only decarbonizes supply (clean power) but also demand, turning compute into a managed utility with sustainability service‑level objectives.

\subsection{Google and Meta: Full-Stack Optimization for Sustainable AI}
\label{subsec:google_meta_sustainable_ai}

Leading technology companies such as Google \cite{elsworth2025measuring} and Meta \cite{wu2022sustainable} have pioneered full-stack optimizations to advance sustainable AI development. Google’s Gemini system \cite{team2023gemini} integrates accelerator-level telemetry, dynamic model routing, and carbon-aware data center scheduling, achieving a 44$\times$ reduction in per-prompt emissions within one year. Meta, formerly Facebook AI, implements end-to-end carbon accounting across all stages of the model lifecycle, from training to inference, while leveraging flash attention, quantization, and sparsity-based techniques to reduce computational demands. Beyond infrastructure optimization, both companies deploy large-scale models for sustainability-critical applications, including climate forecasting, water consumption prediction, and disaster response. These initiatives illustrate how improvements in system-level efficiency, combined with AI-for-good applications, can collectively enable scalable and environmentally responsible AI development.

\subsection{Global AI Applications for Multi-Domain Sustainability}
\label{subsec:global_ai_applications_for_sustainability}

Artificial Intelligence is increasingly recognized as a global enabler of sustainability across multiple domains. In \textit{energy systems}, AI techniques are used for smart grid optimization, dynamic load balancing, and predictive maintenance of renewable infrastructure worldwide \cite{gupta2025comprehensive}. In \textit{materials science}, machine learning accelerates the discovery of photovoltaic and battery materials through high-throughput simulations and generative models, supported by initiatives from the EU and Department of Energy.

In \textit{climate science}, AI enhances the resolution of global climate models via downscaling, improving local policy relevance and disaster preparedness \cite{rolnick2022tackling}. Organizations like NASA and the UN leverage AI-powered early warning systems for floods, wildfires, and hurricanes. Meanwhile, \textit{AI-for-code} research, including AlphaCode \cite{li2022competition}, shows that LLMs can generate energy-efficient software that reduces runtime energy by 10–20\%, promoting greener engineering workflows \cite{solovyeva2025ai}. These global efforts illustrate how AI not only reduces its own footprint but also amplifies sustainability across science, infrastructure, and digital ecosystems.

\vspace{7pt}
\section{Discussion and Future Outlook}
\label{sec:chapter6}

While significant progress has been made in the design of efficient model architectures, the optimization of deployment pipelines, and the development of green AI hardware, many open challenges still remain unresolved. This section provides a reflective discussion on the key limitations and outlines forward-looking perspectives that may define the next phase of green LLM development.




\subsection{Learning without Starting Over: Continual, Incremental, and Federated Training}
\label{subsec:learning_withot_starting_over}

A significant proportion of the carbon footprint associated with LLMs stems from the repeated re-training of monolithic architectures. A more sustainable learning paradigm prioritizes ``learning without starting over." Continual and incremental learning \cite{deng2025multi, yang2024domain, yin2025progressive, tao2020few}, under conditions of model homogeneity, heterogeneity, and modality expansion, should be established as core training paradigms rather than secondary considerations. Promising approaches include: (\textit{i}) elastic parameter partitioning through sparse or adaptive subnetworks that localize model updates and mitigate catastrophic forgetting; (\textit{ii}) retrieval-augmented pretraining and fine-tuning, which externalize knowledge updates into lightweight, mutable indices instead of relying on full-scale weight reconfiguration; and (\textit{iii}) federated and split learning frameworks that enable the exchange of sparsified, differentially private model updates, with provable convergence even under straggler effects and non-IID data distributions. Compiler and runtime systems can further support sustainability by identifying and freezing ``stable" layers, scheduling updates only for ``plastic" regions. Energy-aware Reinforcement Learning from Human Feedback (RLHF) and optimization techniques can penalize computationally intensive update pathways. Collectively, these strategies reduce lifecycle emissions by replacing periodic, resource-intensive model overhauls with targeted, verifiable, and minimal parameter updates.

\subsection{Beyond GPUs: Co-Designing Models with Memory- and Physics-Centric Compute}
\label{subsec:co_designing_models_with_memory_and_physics_centric_compute}

Achieving step-function gains in energy efficiency requires tighter model–hardware fusion and new substrates. Memory-centric compute (\eg, computing near-/in-memory, CNM/CIM) can collapse the von Neumann bottleneck in bandwidth-limited layers (attention/MLP) if models adopt sparsity/low‑rank formats and compilers provide calibration and error‑aware schedules. Neuromorphic/event-driven units are effective for sparse perception and on-device intelligent agents, while photonic interconnects promise ultra‑low‑energy data movement and, in the longer term, enable analog matrix computation. Quantum accelerators remain in early exploration. Critically, models must be co-designed with these emerging hardware platforms in mind: state-space models (\eg, Mamba) that scale linearly with sequence length, and modular routing mechanisms that minimize data movement, \etc. A unified Intermediate Representation (IR) that explicitly captures data movement metrics, such as bytes per operation and data residency levels, combined with autotuners capable of jointly optimizing precision, sparsity, and memory placement, can pave the way for scalable and energy-efficient heterogeneous computing.


\subsection{Edge AI: Efficiency on Constrained Devices}
\label{subsec:edge_ai}

Deploying large models on edge devices remains challenging due to hardware limitations, computational constraints, high energy consumption, and heterogeneous environments. Future advancements will rely on new approaches beyond existing popular techniques like model compression, pruning, and quantization. For example, neuromorphic computing, with brain-inspired architectures, will deliver significantly higher energy efficiency and real-time processing for cognitive tasks through parallel and event-driven operation. Furthermore, quantum neural networks and mobile quantum processing units could revolutionize edge AI by leveraging quantum properties to process massive datasets and optimize decisions at unprecedented speeds, potentially enabling hybrid quantum-classical AI for real-time applications. The synergy with 6G networks will be critical, providing the ultra-low latency and high-speed connectivity required for synchronizing AI across large-scale Internet of Things systems (IoT). Finally, the rise of edge-native AI models optimized for on-device use will reduce cloud dependency and enable more advanced, self-sufficient edge applications.

\subsection{Standardized and Unified Evaluation}
\label{subsec:standardized_and_unified_evaluation}

Equally critical is the lack of standardized, comprehensive benchmarks for evaluating the resource efficiency of large models. Many existing metrics, such as FLOPs, parameter count, or model size, only capture isolated dimensions of efficiency and fail to reflect the complex trade-offs between performance and resource consumption. Moreover, differences in hardware configurations, evaluation protocols, and reporting standards make it difficult to compare methods fairly. This fragmentation hinders progress and obscures the real-world impact of proposed optimizations. A more holistic evaluation framework is needed, one that incorporates not only computational and memory efficiency but also energy consumption, carbon emissions, and lifecycle sustainability. Emerging tools such as CodeCarbon~\cite{gidlund2025green} and experiment-impact-tracker offer promising directions for measuring environmental impact, while performance-efficiency Pareto frontiers provide a principled lens for evaluating trade-offs. Moving forward, the community would greatly benefit from a unified benchmark suite that supports multi-dimensional evaluation across diverse deployment contexts and model sizes.

\vspace{7pt}
\section{Conclusion}
\label{sec:chapter7}

The era of large-scale AI models has brought transformative advances across natural language processing, computer vision, and scientific discovery. However, these advancements entail unprecedented computational and energy demands. This survey explores the green development path of large models, covering efficient architectures, optimized training and inference, hardware-software co-design, and sustainability-driven applications.

We review advances like sparse activation, physics-inspired modeling, dynamic data selection, and parameter-efficient fine-tuning, which reduce resource usage while maintaining performance. On the hardware side, we analyze energy-efficient AI chips, memory-centric computing, and decentralized learning systems that lower environmental impact. We also highlight AI’s growing role in sustainability domains, including remote sensing and green infrastructure.

Looking ahead, we advocate for a shift toward lifecycle-aware model development, standardized efficiency benchmarks, and co-designed algorithm–hardware systems. Embedding environmental responsibility at every layer of AI development will ensure that future models are not only more powerful but also more sustainable and widely accessible.

\acknowledgements{This work was supported in part by the National Natural Science Foundation of China under Grant 62536003, and also in part by the Major Key Project of Pengcheng Laboratory under Grant PCL2025A14.}


\bibliographystyle{cjereport} 
\bibliography{refs}


\end{document}